 \documentclass[final,5p,times,twocolumn]{elsarticle}

\usepackage{cite}
\usepackage{amsmath}
\usepackage{amssymb}
\usepackage{graphicx}
\usepackage{algorithm}
\usepackage{algorithmic}
\usepackage{tikz}
\usepackage{tikz-qtree}
\usepackage{multirow}
\usepackage{rotating}
\usepackage{subfigure}
\usepackage{framed}
\usepackage{color}
\usepackage{colortbl}
\usepackage{longtable}
\usepackage{supertabular}
\usepackage{booktabs}
\usepackage{url}
\definecolor{shadecolor}{rgb}{1,0,0}

\begin{document}

\begin{frontmatter}



\title{A Simple Decomposition-Based Many-Objective Evolutionary Algorithm  with Local Iterative Update}


\author[Yingyu]{Yingyu Zhang}
\ead{zhangyingyu@lcu-cs.com}
\author[Bing]{Bing Zeng}
\ead{zeng.bing.zb@gmail.com}
\address[Yingyu]{School of Computer Science, Liaocheng University, Liaocheng 252059, China.}
\address[Bing]{School of Software Engineering, South China University of Technology, Guangzhou 510006, China.}

\begin{abstract}
Existing studies have shown that the conventional multi-objective evolutionary algorithms (MOEAs) based on decomposition may lose the population diversity when solving some many-objective optimization problems.
In this paper, a simple decomposition-based MOEA with local iterative update (LIU) is proposed.
The LIU strategy has two features that are expected to drive the population to approximate the Pareto Front with good distribution.
One is that only the worst solution in the current neighborhood is swapped out by the newly generated offspring, preventing the population from being occupied by copies of a few individuals.
The other is that its iterative process helps to assign better solutions to subproblems, which is beneficial to make full use of the similarity of solutions to neighboring subproblems and explore local areas in the search space.
In addition, the time complexity of the proposed algorithm is the same as that of MOEA/D, and lower than that of other known MOEAs, since it considers only individuals within the current neighborhood at each update.
The algorithm is compared with several of the best MOEAs on problems chosen from two famous test suites DTLZ and WFG.
Experimental results demonstrate that only a handful of running instances of the algorithm on DTLZ4 lose their population diversity.
What's more, the algorithm wins in most of the test instances in terms of both running time and solution quality, indicating that it is very effective in solving MaOPs.

\end{abstract}


\end{frontmatter}

\section{Introduction}\label{secIntroduction}
A multi-objective optimization problem (MOP) can be formulated as a minimization problem as follows:
\begin{equation}\label{MOP}
\begin{split}
Minimize \quad &F(x)=(f_1(x),f_2(x),...,f_M(x))^T \\
 &Subject \quad to \quad x\in\Omega,
 \end{split}
\end{equation}
where $M\geq 2$ is the number of objective functions, $x$ is a decision vector (or solution),  and $\Omega$ is the feasible set of decision vectors.
A MOP with $M\geq 4$ is often referred to as a many-objective optimization problem (MaOP).
A solution $x$ of Eq.(\ref{MOP}) is said to dominate the other one $y$ ($x\preccurlyeq y$),
if and only if $f_i(x)\leq f_i(y)$ for $i\in(1,...,M)$, and $f_j(x)<f_j(y)$ for at least one index $j\in(1,...,M)$.
A solution $x\in \Omega$ is said to be Pareto-optimal, if there is no solution $y\in \Omega$  such
that $y\preccurlyeq x$. $F(x)$ is then called a Pareto-optimal objective vector.
All the Pareto optimal objective vectors of Eq. (\ref{MOP}) consist of the Pareto Front (PF)\citep{PF}.

Decomposition\citep{Jin2001Adapting, Jaszkiewicz2002On, Ishibuchi1998} has become one of the most famous paradigms for designing multi- or many-objective evolutionary algorithms (MOEAs).
MOEA/D\citep{MOEAD} is a milestone for the decomposition-based MOEAs. In MOEA/D, a MOP is decomposed  into a set of subproblems, by using a set of weight vectors associated with a fitness assignment method such as the penalty-based boundary intersection (PBI) approach.
Since solutions of subproblems associated with the same neighborhood are expected to be similar, they can be updated by a newly generated offspring. In this way, the subproblems are optimized simultaneously.
MOEA/D takes into account both convergence and diversity of the population in a simple framework, and hence have been studied extensively from different points of view \citep{Tam2016, RVEA, Chen2017}.

However, existing studies have shown that the conventional decomposition-based MOEAs have difficulty in solving MaOPs, although they perform well on a wide range of MOPs with two or three
objectives. For example, MOEA/D may lose the population diversity on some problems such as DTLZ4, and the situation gets worse when the number of the objective functions goes up.
To improve the performance of the decomposition-based MOEAs in solving MaOPs, several successful algorithms have been proposed\citep{RVEA, IDBEA, MOEADD}.
One of them is MOEA/DD\citep{MOEADD}.
Like MOEA/D, MOEA/DD employs a set of weight vector associated with the PBI approach to decompose a given MaOP into a set of subproblems and optimize them simultaneously.
To improve the performance, MOEA/DD incorporates a hybrid update strategy based on decomposition and dominance, in which dominance is used as a ranking method to help globally select the worst solution from the whole population.
Experimental results have shown that MOEA/DD outperforms MOEA/D and several of the best MOEAs, i.e., NSGA-III\citep{NSGAIII}, GrEA\citep{GrEA} and  HyPE\citep{HYPE}, indicating that its update strategy is effective to avoid the loss of the population diversity.

 The update strategy adopted by MOEA/DD has two different features from MOEA/D. One is that, it is a global update strategy that considers all solutions in the whole population at each update, while MOEA/D considers only solutions within the current neighborhood. Therefore, it takes more time for MOEA/DD to update the population than MOEA/D.
The other is that, only the worst solution of the whole population is replaced out by the newly generated offspring. But in MOEA/D, all of the solutions within the current neighborhood that worse than the newly generated offspring will be replaced. That is the reason why the population of MOEA/D can be occupied by copies of several solution, and loses its diversity. Li and Zhang\citep{MOEAD-DE} have noticed this problem and modified the update strategy to replace at most two solutions within the current neighborhood.
This slows down the loss of the population diversity, but can not stop it.

In this paper, a simple local iterative update (LIU) strategy is proposed and incorporated into a simplified decomposition-based MOEA framework. The resulting algorithm is hence denoted as MOEA/D-LIU. It has the same computational complexity as MOEA/D, but shows competitive performance compared with several of the best MOEAs.
The proposed LIU strategy has the following features:
\begin{enumerate}
  \item It considers only solutions within the current neighborhood at each update. Therefore, the time complexity of our algorithm is the same as that of MOEA/D, and lower than that of other existing MOEAs, such as NSGA-II, MOEA/DD and RVEA\citep{RVEA}, etc. Especially, almost all of the actual running times of our algorithm in our experiments are lower than those of MOEA/D.
  \item Only the worst solution on the iterative path is replaced by the newly generated offspring. This further slows down the loss of the population diversity. And only a handful of running instances of our algorithm on DTLZ4 lose their population diversity in our experiments.
  \item Its iterative process helps to assign better solutions to subproblems. One of the biggest advantages of the decomposition-based MOEAs is that they considers similarity between solutions of neighboring subproblems and each subproblem is optimized by using information from its neighboring subproblems. By assigning solutions to more appropriate subproblems, the LIU strategy can make full use of this advantage. Overall, MOEA/D-LIU outperforms MOEA/D, MOEA/DD and GrEA in most of the comparisons of DTLZ1 to DTLZ4 and WFG1 to WFG9 in terms of convergence and diversity.
\end{enumerate}

The rest of the paper is organized as follows. In Section II, we provide some preliminaries used in MOEA/D-LIU.
The algorithm MOEA/D-LIU is proposed in Section III, including the initialization procedure,  the reproduction procedure and the LIU procedure.
Some discussions about similarities and differences between the algorithm and MOEA/D are also made.
In Section IV, experimental data obtained by MOEA/D-LIU on DTLZ1 to DTLZ4 and WFG1 to WFG9 are compared with those obtained by several other MOEAs, i.e., MOEA/D, MOEA/DD and GrEA.
The paper is concluded in Section V.
\section{Prelimilaries}\label{secPrelimilaries}
\subsection{Systematic Sampling Approach}\label{subSectionSSA}
A weight vector generation method is needed in a decomposition-based MOEA. We use the systematic sampling approach proposed by Das and Dennis\citep{SystematicApproach}
to generate weight vectors in our algorithm presented later in Section \ref{secProposedAlgorithm}.
In this approach, weight vectors are sampled from a unit simplex.
Let $w=(w_1,...,w_M)^T$ is a given weight vector, $w_j(1\leqslant j\leqslant M)$ is the $jth$ component of $w$,
$\delta_j$ is the uniform spacing between two consecutive $w_j$ values, and $1/\delta_j$ is an integer.
The possible values of $w_j$ are sampled from $\{0,\delta_j,...,K_j\delta_j\}$, where $K_j=(1-\sum_{i=1}^{j-1}w_i)/\delta_j$.
In a special case, all $\delta_j$ are equal to $\delta$.
To generate a weight vector, the systematic sampling approach starts with sampling from $\{0,\delta,2\delta,...,1\}$ to obtain the first component $w_1$, and
then from $\{0,\delta,2\delta,...,K_2\delta\}$ to get the second component $w_2$  and so forth, until the $Mth$ component $w_M$
is generated. Repeat such a process, until a total of
\begin{equation}\label{nWeightVectors}
N(D,M)=\left(
\begin{array}{c}
  D+M-1 \\
  M-1
\end{array}
\right)
\end{equation}
different weight vectors  are generated, where $D > 0$ is the number of divisions considered along each objective coordinate.

However, a MOEA with a large D would generate too much weight vectors and this in turn would add more computational burden to the MOEA, and a small D would  be harmful  to the population diversity.
Therefore, a two-layer weight vector generation method\citep{NSGAIII, MOEADD} is adopted in our algorithm for test instances with $M\geq8$.
In the two-layer weight vector generation method, a set of $N_1$ weight vectors in the boundary layer and a set of $N_2$ weight vectors in the inside layer are generated in advance,
according to the systematic sampling approach described above.
The coordinates of each weight vector in the inside layer are then shrunk by a coordinate transformation as
\begin{equation}
v_{i}=\frac{1-\tau}{M}+\tau\times w_{i},
\end{equation}
where $w_{i}$ is the ith component of the weight vector $w$ in the  inside layer, and $\tau\in [0,1]$ is a shrinkage factor set as
$\tau=0.5$ in \citep{NSGAIII} and \citep{MOEADD}.
Finally, the two sets are combined to form the final set of weight vectors.
Denote the number of the weight vectors generated in the boundary layer and the inside layer as D1 and D2 respectively.
Then, the total number of the weight vectors generated by the two-layer weight vector generation method is
$N(D1,M)+N(D2,M)$.

\subsection{PBI approach}\label{subSectionPBI}
The PBI approach is adopted in our algorithm as a fitness assignment method to decompose a problem into subproblems, and serves as a criterion for comparing solutions\citep{MOEAD}.
In the PBI criterion, a solution $x$ is considered to be better than the other one $y$ when $PBI(x)<PBI(y)$.
To calculate the PBI value of a solution $x$, the objective values of it are needed to be transformed beforehand as:
\begin{equation}\label{objectiveTranformation}
f_i'(x)=f_i(x)-z^*_i,
\end{equation}
where $z^*=(z^*_1, z^*_2,..., z^*_M)^T$ is the ideal point, and $f_i(x)$ is the ith objective value of $x$.

The PBI value is then computed as\citep{MOEAD}:
\begin{equation}\label{PBI}
PBI(x)=PBI(x,w)=d_1+\theta d_2
\end{equation}
with
\begin{equation}\label{twoDists}
\begin{split}
&d_1=d_1(x,w)=\frac{\left\|(F'(x))^{T}w\right\|}{\|w\|},  \\
&d_2=d_2(x,w)=\left\|F'(x)-d_1\frac{w}{\|w\|}\right\|
\end{split}
\end{equation}
that are used as measures for convergence and diversity of the population, respectively,
where $w$ is a given weight vector, $\theta$ is a user-defined constant penalty parameter, and $F'(x)=(f_1'(x), f_2'(x),..., f_M'(x))^T$.

\subsection{Objective Normalization}
Objective normalization has been proven to be effective\citep{Cvetkovic2002MOP, Osiadacz1989MOP, Coit1998Genetic} for a MOEA to solve MaOPs with disparately scaled objectives such as ZDT3 and WFG1 to WFG9. In this paper, we adopt a simple  normalization method\citep{MOEAD} that transforms each objective according to the following form:
\begin{equation}\label{objectiveNormalization}
f''_i(x)=\frac{f_i(x)-z^{*}_{i}}{z_i^{nad}-z^{*}_{i}},
\end{equation}
where $z_i^{nad}=\max\{f_i(x)|x\in PS\}$, and $z^{nad}=(z_1^{nad},z_2^{nad},...,z_M^{nad})^T$ is the nadir point.
Eq. (\ref{twoDists}) can hence be rewritten as
\begin{equation}\label{twoDists2}
\begin{split}
&d_1=\frac{\left\|(F''(x))^{T}w\right\|}{\|w\|}\\
&d_2=\left\|F''(x)-d_1\frac{w}{\|w\|}\right\|,
\end{split}
\end{equation}
where $F''(x)=(f''_1(x), f''_2(x),..., f''_M(x))^T$.

However, it is generally not easy to obtain $z^{*}$ and $z^{nad}$ in advance,
therefore we replace $z^{*}_i$ and $z_i^{nad}$ with the minimum and maximum value, respectively, of the ith objective that have been found so far.
To be specific, $z^{*}$ and $z^{nad}$ are initialized by the objective values of the initial population in the initialization phase. Thereafter, both of them are updated by every newly generated offspring.

\section{Proposed Algorithm:MOEA/D-LIU}\label{secProposedAlgorithm}
\subsection{Algorithm Framework}\label{subSectionAlgFramework}
The proposed algorithm MOEA/D-LIU is implemented in a simplified version of the original decomposition-based MOEA framework that is presented in Algorithm \ref{algFramework}.
At each generation, the algorithm traverses  $N$ weight vectors, generates an offspring for each weight vector by using several conventional  reproduction operators, i.e., the selection operator, the simulated binary crossover (SBX) operator\citep{SBX} and the polynomial mutation (PM) operator\citep{PM}, and updates the population with the offspring in the LIU procedure.

\begin{algorithm}
\caption{General Framework of MOEA/D-LIU}
\label{algFramework}
\begin{algorithmic}[1]
\REQUIRE Maximum Number of Generations:G.
\ENSURE  Final Population.
\STATE   Initialization Procedure;
\FOR{$t=1$ to $G$}
    \FOR{$i=1$ to $N$}
    \STATE Reproduction Procedure;
    \STATE LIU procedure;
    \ENDFOR
\ENDFOR
\end{algorithmic}
\end{algorithm}

\subsection{Initialization Procedure}
The initialization procedure includes four steps.
In the first step, a set of uniformly distributed weight vectors are generated using the systematic sampling approach described in Section \ref{subSectionSSA}.

In the second step, the neighborhood of each weight vector is generated by calculating the included angles between the weight vector and other weight vectors.
The included angle between two weight vectors $w$ and $w'$ can be calculated as:
\begin{equation}\label{angle}
\tan\theta=\frac{d_2'}{d_1'},
\end{equation}
where $d_1'$ and $d_2'$ are variants of $d_1$ and $d_2$ in Eq. (\ref{twoDists}), respectively.
They can be calculated as
\begin{equation}\label{twoDists3}
d_1'=\frac{\left\|w^{T}w'\right\|}{\|w'\|}, \quad d_2'=\left\|w-d_1\frac{w'}{\|w'\|}\right\|.
\end{equation}
A set of T weight vectors that have the minimum angles to the weight vector forms the neighborhood.

In the third step,  $N$ individuals are initialized randomly one by one.
To initialize the ith individual is to generate n random numbers $x_j\in[L_j,U_j]\ (1\leq j\leq n)$ for the individual as its decision variables, and evaluate it.
$L_j$ and $U_j$ are the lower and upper bounds of the jth decision variable, respectively.
The individual is then attached to the ith weight vector. In other words, the ith randomly generated individual represents the initial solution $x=(x_1, x_2,..., x_n)$ of the ith subproblem.

In the fourth step, the ideal point is initialized as
\begin{equation}
z_i^{*}=\min_{x\in P}f_i(x),
\end{equation}
where, $P$ is currently the initial population, $f_i(x)$ is the ith objective of an individual x.
Similarly, the nadir point is initialized as
\begin{equation}
z_i^{nad}=\max_{x\in P}f_i(x).
\end{equation}

\subsection{Reproduction Procedure}
The reproduction procedure can be described as follows.
Firstly, two individuals are selected as the mating parents. The first individual is assigned to be the individual attached to the current weight vector. A random number $r$ between 0 and 1 is then generated. If $r$ is less than a given selection probability $p_s$,
then chooses the second individual from the neighborhood of the current weight vector, or else randomly chooses the second individual from the whole population.

Secondly, the SBX operator is applied on the two parents to generate two intermediate individuals.
Denote the two parents as $p_1=(p_{1,1}, ..., p_{1,n})$, and $p_2=(p_{2,1}, ..., p_{2,n})$, respectively.  The two intermediate individuals $\overline{c}_1=(\overline{c}_{1,1}, ..., \overline{c}_{1,n})$ and $\overline{c}_2=(\overline{c}_{2,1}, ..., \overline{c}_{2,n})$ are calculated as follows:
\begin{equation}
\begin{split}
\overline{c}_{1,i}=0.5\left[(1+\beta_i)p_{1,i}+(1-\beta_i)p_{2,i}\right],\\
\overline{c}_{2,i}=0.5\left[(1-\beta_i)p_{1,i}+(1+\beta_i)p_{2,i}\right],
\end{split}
\end{equation}
with
\begin{equation}
\beta_i=
\left\{
\begin{aligned}
&(2u_i)^{\frac{1}{\eta_c+1}},\ & if \  u_i\leq 0.5;\\
&\left(\frac{1}{2(1-u_i)}\right)^\frac{1}{\eta_c+1}, \ & otherwise,
\end{aligned}
\right.
\end{equation}
where $u_i\in [0,1]$ is a random number, and $\eta_c$ is the distribution index.
Notice that, if both of the two individuals are preserved for the following steps,
then the number of  individuals evaluated at each generation will be twice as many as the population size.
However, the number of individuals evaluated at each generation in the MOEAs to be compared with
is the same as the population size. Therefore, one of the two intermediate individuals is abandoned at random for the sake of fairness.

Thirdly, the PM operator is applied on the preserved intermediate individual to generate an
offspring $c$, which will be evaluated and used to update the current population in the following LIU procedure.The PM operator generates an offspring $c=(c_1, ..., c_n)$ from the preserved intermediate individuals $\overline{c}=(\overline{c}_1, ..., \overline{c}_n)$ in the following way:
\begin{equation}
c_i=\overline{c}_i+(U_i-L_i)\delta_i,
\end{equation}
with
\begin{equation}
\delta_i=
\left\{
\begin{aligned}
&(2u_i)^{\frac{1}{\eta_m+1}},\ & if \  u_i\leq 0.5;\\
&1-[2(1-u_i)]^{\frac{1}{\eta_m+1}}, \ & otherwise,
\end{aligned}
\right.
\end{equation}
where $u_i\in [0,1]$ is a random number, and $\eta_m$ is the distribution index.

Fourthly, update the ideal point. If the ith objective value $f_i(c)$ of the offspring is less than $z_i^*$, replace $z_i^*$ with $f_i(c)$.

Finally, update the nadir point. If the ith objective value $f_i(c)$ of the offspring is larger than $z_i^{nad}$, replace $z_i^{nad}$ with $f_i(c)$.

\subsection{LIU Strategy}
As it can be seen from Algorithm \ref{algLIUStrategy}, the  LIU  strategy is direct and very simple. It traverses  T weight vectors within the current neighborhood, compare the individual associated with each weight vector and the individual stored in the $c$ to judge whether to swap them.
Specifically, at each iterative step, the jth individual $\mathcal{N}[i][j]$, i.e., the individual associated with the jth subproblem (weight vector) in the current neighborhood $\mathcal{N}[i]$, is compared with the individual stored in $c$.
The swapping operation occurs only when the individual stored in $c$ is better than $\mathcal{N}[i][j]$.
In other words, a swapping operation assigns a better solution to the subproblem associated with the current weight vector (jth weight vector in the current neighborhood).
The last individual stored in $c$ is considered as the worst individual in the current neighborhood and abandoned.

\begin{algorithm}
\caption{The LIU Strategy}
\label{algLIUStrategy}
\begin{algorithmic}[1]
\REQUIRE The  offspring $c$ and the current neighborhood $\mathcal{N}[i]$.
\FOR{$j=1$ to $T$}
\IF{$PBI(c)<PBI(\mathcal{N}[i][j])$}
\STATE $Swap(c,\mathcal{N}[i][j])$;
\ENDIF
\ENDFOR
\end{algorithmic}
\end{algorithm}

The LIU strategy is simple but has two features that can help the population evolve more effectively.
One is that, only the worst solution in the current neighborhood is replaced out of the population, preventing the population from being occupied by copies of a few individuals.
The other is that, its iterative process helps to arrange individuals, that is, to assigns better solutions to subproblems within the current neighborhood.

\subsection{Time Complexity}
As it is mentioned above, the  LIU  strategy traverses T weight vectors within the current neighborhood,
compare the individual associated with each weight vector and the individual stored in the $c$ to judge whether to swap them.
Therefore, it takes $O(MT)$ times of floating-point calculations for the LIU strategy  to update the population, where M is the number of the objective functions and $T$ is the neighborhood size.
The time complexity of the proposed algorithm at each generation is hence $O(MNT)$, which is the same as that of MOEA/D, and lower than that of other known MOEAs, such as MOEA/DD, NSGA-II and RVEA\citep{RVEA}, etc.
It is shown in Section \ref{secPerformanceComparisons} that, the actual running times of the algorithm on almost all instances of DTLZ1 to DTLZ4 and WFG1 to WFG4 are less than those of MOEA/D and MOEA/DD, indicating that the algorithm is computationally efficient.

\subsection{Discussion}\label{discussion}
As we can see, MOEA/D-LIU is a variant of MOEA/D. Like MOEA/D, our algorithm uses a set of weight vectors associated with a fitness assignment method to decompose a given MaOP into subproblems and optimize them simultaneously. In addition, Both MOEA/D-LIU and MOEA/D consider only the individuals within the current neighborhood at each update, which makes the time complexity of them the lowest among known MOEAs. But they differ in the following three aspects:
\begin{enumerate}
  \item MOEA/D-LIU adopts an angle-based approach to identify the neighborhood of a subproblem, while MOEA/D uses a method based on the Euclidean distance between reference points.
  \item In MOEA/D, all individuals in the current neighborhood that are worse than the newly generated offspring  will be replaced.
      Whereas in MOEA/D-LIU, only the worst solution in the current neighborhood will be swapped out and abandoned. This can effectively prevent the current population from being occupied by copies of a few individuals, and losing its diversity.
  \item MOEA/D doesn't change the position of a preserved individual throughout the update process. On the contrary, the iterative process of the LIU strategy used by MOEA/D-LIU  helps to arrange individuals within the current population. In other words, the LIU strategy is designed not only to preserve better solutions to the next generation, but also to assign the right solution to each subproblem. Meanwhile, solutions of neighboring subproblems of a continuous MaOP are assumed to be similar, they are kept similar in this way. This is beneficial to make full use of the advantages of decomposition.
      For example, the similarity of solutions of neighboring subproblems can help the neighborhood selection operation to explore local areas more effectively, which is discussed a little further in Section \ref{parameterSensitivityStudies}.
\end{enumerate}

Based on the above points, the LIU strategy is expected to drive the population to approximate the PF quickly with good distribution.

\section{Experimental Results}\label{secExperimentalResults}
The proposed algorithm MOEA/D-LIU is implemented\footnote{The \  source \   code \  of \   MOEA/D-LIU \  can \  be \  downloaded \  from:\\ \url{https://share.weiyun.com/59kVbCt}. The source code of jMetal 5.4 can be downloaded from: \url{http://jmetal.github.io/jMetal/}.} in the framework of jMetal 5.4\citep{jMetal2011,jMetal2015} and run 20 times independently for each instance on Intel(R) Core(TM) i7- 4790 CPU. In this section, the running times of the algorithm are compared with those obtained by MOEA/D and MOEA/DD, and the experimental results are compared with those appeared in \citep{MOEADD}. We also conduct a study on the parameter sensitivity of the algorithm.

\subsection{Performance Metrics}\label{performanceMetris}
The following two performance metrics are employed to measure the convergence and diversity of the algorithms to be compared.
\subsubsection{Inverted Generational Distance(IGD)}
Let $S$ be a solution set of a MOEA on a given MaOP, and $R$ be a set of uniformly distributed representative points of the PF.
The IGD value of $S$ relative to $R$ can be calculated as\citep{IGD}
\begin{equation}
IGD(S,R)=\frac{\sum_{r\in R}d(r,S)}{|R|}
\end{equation}
where $d(r,S)$ is the minimum Euclidean distance between $r$ and the points in $S$, and $|R|$ is the cardinality of $R$. Note that, the points in $R$ should be well distributed and $|R|$ should be large enough to ensure that the points in $R$ could represent the PF very well. This guarantees that the IGD value of $S$ is able to measure the convergence and diversity of the solution set. The lower the IGD value of $S$, the better its quality\citep{MOEADD}.

\subsubsection{HyperVolume(HV)}
The HV value of a given solution set $S$ is defined as\citep{HV}
\begin{small}
\begin{equation}
HV(S)=vol\left( \bigcup_{x\in S}\left[ f_1(x),z_1 \right]\times \ldots \times\left[ f_M(x),z_M \right]\right),
\end{equation}
\end{small}
where $vol(\cdot)$ is the Lebesgue measure, and $z^r=(z_1,\ldots,z_M)^T$ is a given reference point. As it can be seen that the HV value of $S$ is a measure of the size of the objective space dominated by the solutions in $S$ and bounded by $z^r$.

As with \citep{MOEADD}, an algorithm based on Monte Carlo sampling proposed in \citep{HYPE}  is applied to compute the approximate HV values for 15-objective test instances, and the WFG algorithm \citep{WFGalgorithm} is adopted to compute the exact HV values for other test instances for the convenience of comparison. In addition, all the HV values are normalized to $[0,1]$ by dividing $\prod_{i=1}^{M}z_i$.

\subsection{Benchmark Problems}
\subsubsection{DTLZ test suite}
Problems DTLZ1 to DTLZ4 from the DTLZ test suite proposed by Deb et al\citep{DTLZ}  are chosen for our experimental studies in the first place.
One can refer to \citep{DTLZ} to find their definitions.
Here, we only summarize some of their features.
\begin{itemize}
  \item DTLZ1:The global PF of this problem is the linear hyper-plane $\sum_{i=1}^{M}f_i=0.5$. And the search space contains $(11^k-1)$  local PFs that can hinder a MOEA to converge to the hyper-plane.
  \item DTLZ2:The global PF of this problem satisfies $\sum_{i}^{M}f_i^2=1$. Previous studies have shown that this problem is easier to be solved by existing MOEAs, such as NSGA-III, MOEADD, etc., than DTLZ1, DTLZ3 and DTLZ4.
  \item DTLZ3:This problem has the same PF shape as DTLZ2, and introduces $(3^k-1)$ local PFs that are parallel to the global PF. A MOEA can get stuck at any of these local PFs before converging to the global PF. Therefore, this problem can be used to investigate a MOEA's ability to converge to the global PF.
  \item DTLZ4:This problem also has the same PF shape as DTLZ2, and can be obtained by modifying DTLZ2 with a different meta-variable mapping,  which is expected to introduce a biased density of solutions in the search space.  Therefore, it can be used to investigate a MOEA's ability to maintain a good distribution of solutions.
\end{itemize}

To calculate the IGD value of a result set $S$ obtained by a MOEA on a MaOP, a set $R$ of representative points of the PF needs to be given in advance.
For problems DTLZ1 to DTLZ4,  we take the set of the intersecting points of the weight vectors
and the PF surface as $R$.
Let $f^*=(f_{1}^*,...,f_{M}^*) $ be the intersecting point of a weight vector $w=(w_1,...,w_M)^T$ and the PF surface.
Then $f_i^*$ can be computed as\citep{MOEADD}
\begin{equation}
f_i^*=0.5\times\frac{w_i}{\sum_{j=1}^{M}w_j}
\end{equation}
for DTLZ1, and
\begin{equation}
f_i^*=\frac{w_i}{\sqrt{\sum_{j=1}^{M}w_j}}
\end{equation}
for DTLZ2, DTLZ3 and DTLZ4.

\subsubsection{WFG test suite}
This test suite allows test problem designers to construct scalable test problems with any number of objectives,
in which features such as modality and separability can be customized as required.
As discussed in \citep{WFGProblems,WFG},
it exceeds the functionality of the DTLZ test suite.
In particular, one can construct non-separable problems, deceptive problems,
truly degenerative problems, mixed shape PF problems,
problems scalable in the number of position-related parameters,
and problems with dependencies between position- and distance-related parameters as well with
the WFG test suite.

In \citep{WFG}, several scalable problems, i.e., WFG1 to WFG9,
are suggested for MOEA designers to test their algorithms,
which can be described as follows.
\begin{equation}
\begin{split}
Minimize \quad F(X)&=(f_1(X),...,f_M(X))\\
f_i(X)&=x_M+2ih_i(x_1,...,x_{M-1}) \\
X&=(x_1,...,x_M)^T
\end{split}
\end{equation}
where $h_i$ is a problem-dependent shape function  determining the geometry of the fitness
space, and $X$ is derived from a vector of working parameters $Z=(z_1,...,z_n)^T, z_i\in [0,2i]$ , by employing four problem-dependent transformation functions $t_1$, $t_2$, $t_3$  and  $t_4$.
Transformation functions must be designed carefully such that the underlying PF remains intact with a relatively easy to determine Pareto optimal set.
The WFG Toolkit provides a series of predefined shape and transformation functions to help ensure this is the case.
One can refer to  \citep{WFGProblems,WFG} to see their definitions.
Let
\begin{equation}
\begin{split}
Z''&=(z''_1,...,z''_m)^T
=t_4(t_3 (t_2 (t_1(Z'))))\\
Z'&=(z_1/2,...,z_n/2n)^T.
\end{split}
\end{equation}
Then $x_i=z''_i(z''_i-0.5)+0.5$ for problem WFG3, whereas $X=Z''$ for problems WFG1, WFG2 and WFG4 to WFG9.

The features of WFG1 to WFG9 can be summarized as follows.
\begin{itemize}
  \item WFG1:A separable and unimodal problem with a biased PF and a convex and mixed geometry.
  \item WFG2:A non-separable problem with a convex and disconnected geometry, i.e., the PF of WFG2 is composed of several disconnected convex segments. And all of its objectives but $f_M$ are unimodal.
  \item WFG3:A non-separable and unimodal problem with a linear and degenerate PF shape, which can be seen as a connected version of WFG2.
  \item WFG4:A separable and multi-modal problem with large "hill sizes", and a concave geometry.
  \item WFG5:A separable and deceptive problem with a concave geometry.
  \item WFG6:A nonseparable and unimodal problem with a concave geometry.
  \item WFG7:A separable and unimodal problem with parameter dependency, and a concave geometry.
  \item WFG8:A nonseparable and unimodal problem with parameter dependency, and a concave geometry.
  \item WFG9:A nonseparable, deceptive and unimodal problem with parameter dependency, and a concave geometry.
\end{itemize}

As it can be seen from above, WFG1 and WFG7 are both separable and unimodal,
and WFG8 and WFG9 have nonseparable property,
but the parameter dependency of WFG8 is much harder than that of WFG9.
In addition, the deceptiveness of WFG5 is more difficult than that of WFG9,
since WFG9 is only deceptive on its position parameters.
However, when it comes to the nonseparable reduction, WFG6 and WFG9 are more difficult than  WFG2 and WFG3.
Meanwhile,problems WFG4 to WFG9 share the same PF shape in the objective space,
which is a part of a hyper-ellipse with radii $r_i = 2i$, where $i\in\{1,...,M\}$.

\subsection{Parameter Settings}\label{parameterSettings}

\begin{table}
\caption{Divisions and Population Sizes}
\begin{center}\label{nDivisions}
\begin{tabular}{|c|c|c|c|}
\hline
M&D1&D2&Population Size\\
\hline
3&12&-&91\\
5&6&-&210\\
8&3&2&156\\
10&3&2&275\\
15&2&1&135\\
\hline
\end{tabular}
\end{center}
\end{table}

\begin{table}
\caption{Number OF Generations}
\begin{center}
\begin{tabular}{|c|c|c|c|c|c|}

  \hline
  Instance& $M=3$ & $M=5$ & $M=8$ & $M=10$ & $M=15$ \\
  \hline
 DTLZ1 & 400 & 600 & 750 & 1000 & 1500 \\
  DTLZ2 & 250 & 350 & 500 & 750 & 1000 \\
  DTLZ3 & 1000 & 1000 & 1000 & 1500 & 2000 \\
  DTLZ4 & 600 & 1000 & 1250 & 2000 & 3000 \\
  \hline
\end{tabular}\label{nGens}
\end{center}
\end{table}

The algorithm  does not require any new parameters except some parameters in the conventional MOEAs, such as the population size, the selection probability, and so on, which can be listed as follows:
\begin{itemize}
  \item Settings for Crossover Operator:The crossover probability is  $p_c=1.0$ and the distribution index is $\eta_c=20$.
  \item Settings for Mutation Operator:The mutation probability is  $p_m=0.5/n$, and the distribution index is  $\eta_m=20$.
  \item Population Size:The population size of MOEA/D-LIU is the same as the number of the weight vectors. For 3- and 5-objective instances, the weight vectors are generated by the original systematic sampling approach, and  for 8-, 10- and 15-objective instances, the two-layer weight vector generation method is applied. In both cases, the number of divisions along each coordinate determines the number of the weight vectors as discussed previously.  The number of divisions and the population size  are summarized in Table \ref{nDivisions}.
  \item Number of Runs:The algorithm is independently run 20 times on each test instance.
  \item Number of Generations: MOEA/D-LIU stopped at a predefined number of generations. The number of generations for DTLZ1 to DTLZ4 is listed in Table \ref{nGens}, and the number of generations for all the instances of WFG1 to WFG9 is set to 3000.
  \item Penalty Parameter in PBI: $ \theta= 5.0$.
  \item Neighborhood Size: $T = 30$.
  \item Selection Probability: The probability of selecting two mating individuals from the current neighborhood is  $p_s = 0.9$.
  \item Settings for DTLZ1 to DTLZ4:As in papers\citep{MOEADD, IDBEA},
  the number of the objectives are set as $M \in \{3,5,8,10,15\}$ for comparative purpose.
  And the number of the decision variables is  $n = M + r-1$, where $r = 5$ for DTLZ1, and $r = 10$ for DTLZ2, DTLZ3 and DTLZ4.
  To calculate the HV value  we set the reference point to $(1,...,1)^T$ for DTLZ1,  and $(2,...,2)^T$ for DTLZ2 to DTLZ4.
  \item Settings for WFG1 to WFG9:
  The number of the decision variables is  $n = k + l$,
  where   $k = 2\times(M-1)$  is the position-related variable and $l = 20$ is the distance-related variable.
  To calculate the HV values for problems WFG1 to WFG9, the reference point is $(3,...,2M+1)^T$.
\end{itemize}

For the sake of fairness, the population size, the number of runs and the number of generations are the same as  other algorithms for comparison.

\subsection{Performance Comparisons on DTLZ1 to DTLZ4}\label{secPerformanceComparisons}

\begin{table*}[!htbp]
  \centering
  \caption{Running times (in milliseconds) of MOEA/D-LIU, MOEA/D and MOEA/DD on the instances of problems DTLZ1 to DTLZ4. }
  \resizebox{\linewidth}{!}{
    \begin{tabular}{|c|c|c|c|c|c|c|c|c|}
    \toprule
    Problem & M     & MOEA/D-LIU & MOEA/D & MOEA/DD & Problem & MOEA/GLU & MOEA/D & MOEA/DD \\
    \midrule
    \multirow{5}[10]{*}{DTLZ1} & 3     & \cellcolor[rgb]{ .851,  .851,  .851}635  & 711   & 1012  & \multirow{5}[10]{*}{DTLZ2} & \cellcolor[rgb]{ .851,  .851,  .851}481  & 598   & 698  \\
\cmidrule{2-5}\cmidrule{7-9}          & 5     & \cellcolor[rgb]{ .851,  .851,  .851}2871  & 3534  & 10374  &       & \cellcolor[rgb]{ .851,  .851,  .851}1951  & 2515  & 7053  \\
\cmidrule{2-5}\cmidrule{7-9}          & 8     & \cellcolor[rgb]{ .851,  .851,  .851}3481  & 3898  & 9733  &       & \cellcolor[rgb]{ .851,  .851,  .851}2658  & 3095  & 7237  \\
\cmidrule{2-5}\cmidrule{7-9}          & 10    & \cellcolor[rgb]{ .851,  .851,  .851}9607  & 12976  & 49295  &       & \cellcolor[rgb]{ .851,  .851,  .851}7545  & 8925  & 42291  \\
\cmidrule{2-5}\cmidrule{7-9}          & 15    & \cellcolor[rgb]{ .851,  .851,  .851}9350  & 9394  & 25473  &       & \cellcolor[rgb]{ .851,  .851,  .851}7341  & 8363  & 18098  \\
    \midrule
    \multirow{5}[10]{*}{DTLZ3} & 3     & \cellcolor[rgb]{ .851,  .851,  .851}2031  & 2402  & 2667  & \multirow{5}[10]{*}{DTLZ4} & \cellcolor[rgb]{ .851,  .851,  .851}1266  & 1370  & 1661  \\
\cmidrule{2-5}\cmidrule{7-9}          & 5     & \cellcolor[rgb]{ .851,  .851,  .851}5841  & 7537  & 17811  &       & \cellcolor[rgb]{ .851,  .851,  .851}6109  & 7020  & 20077  \\
\cmidrule{2-5}\cmidrule{7-9}          & 8     & \cellcolor[rgb]{ .851,  .851,  .851}5410  & 5641  & 12958  &       & \cellcolor[rgb]{ .851,  .851,  .851}7417  & 8290  & 16981  \\
\cmidrule{2-5}\cmidrule{7-9}          & 10    & \cellcolor[rgb]{ .851,  .851,  .851}16843  & 19364  & 48936  &       & \cellcolor[rgb]{ .851,  .851,  .851}22181  & 23097  & 105682  \\
\cmidrule{2-5}\cmidrule{7-9}          & 15    & 13736  & \cellcolor[rgb]{ .851,  .851,  .851}10795  & 21037  &       & 24949  & \cellcolor[rgb]{ .851,  .851,  .851}24309  & 57763  \\
    \bottomrule
    \end{tabular}}%
  \label{tab:AVGRuntimesOfDTLZ}%
\end{table*}%

\begin{figure*}[!htbp]
\begin{minipage}[t]{0.5\linewidth}
    \centering
    \includegraphics[width=\textwidth]{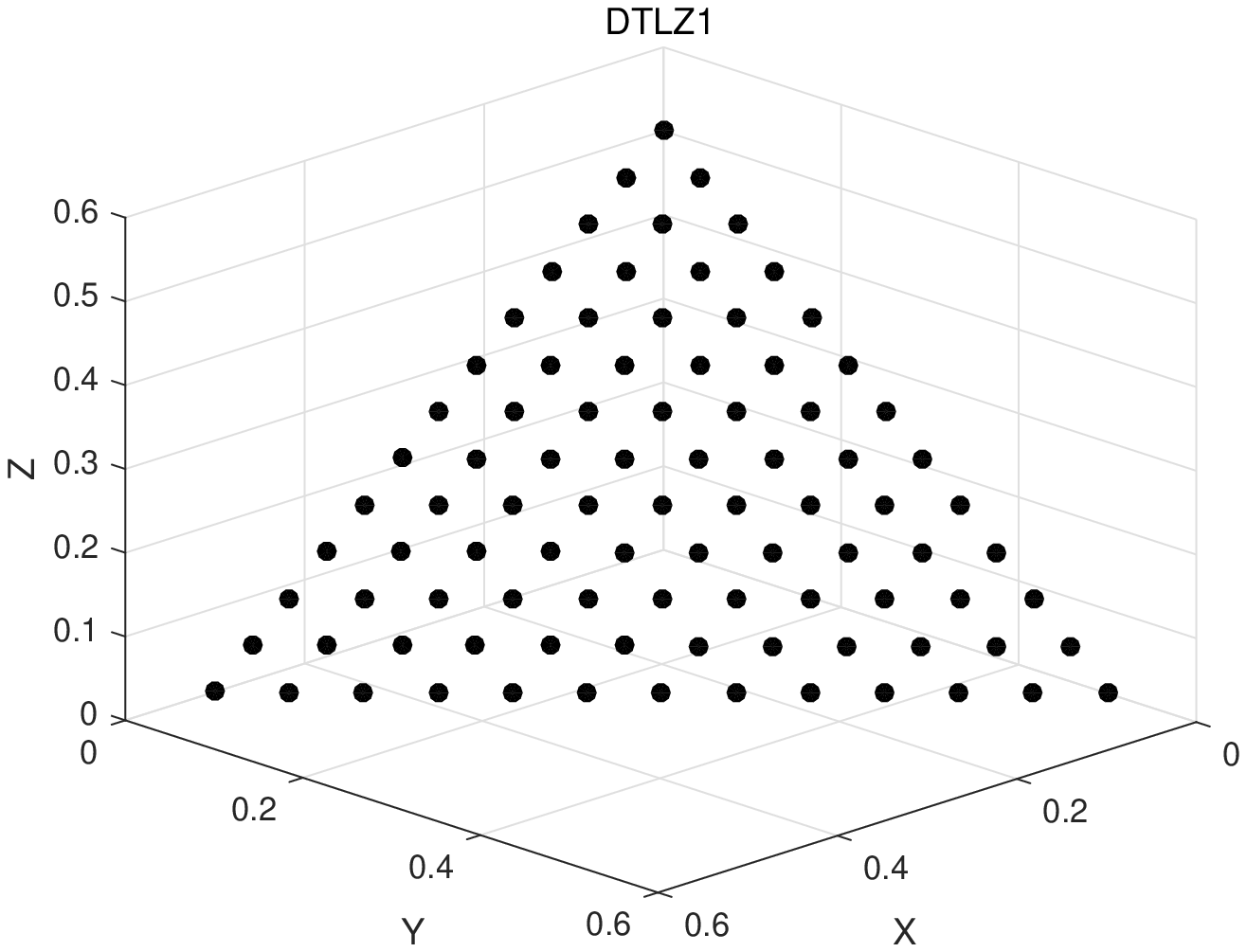}
\end{minipage}
\begin{minipage}[t]{0.5\linewidth}
    \centering
    \includegraphics[width=\textwidth]{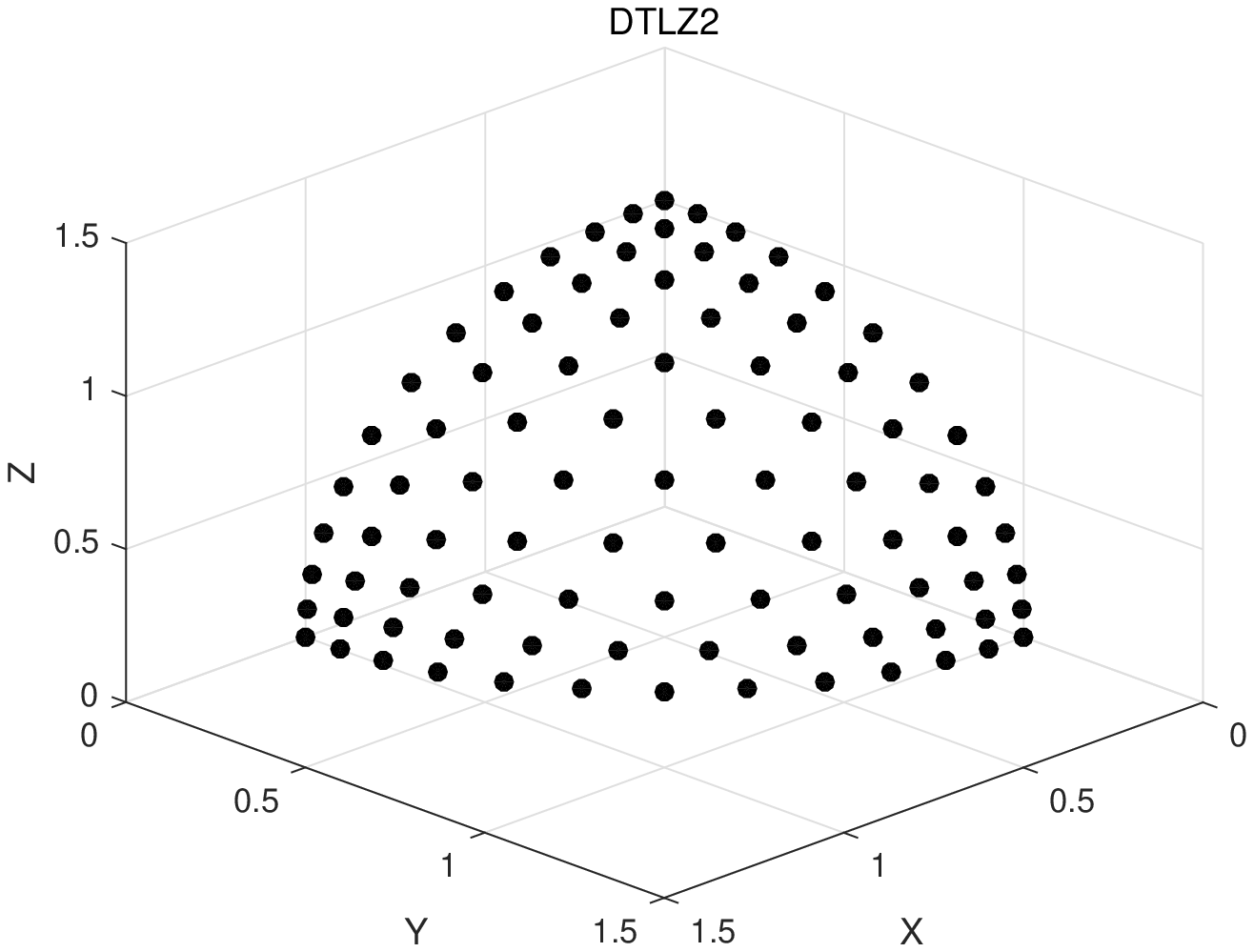}
\end{minipage}
\begin{minipage}[t]{0.5\linewidth}
    \centering
    \includegraphics[width=\textwidth]{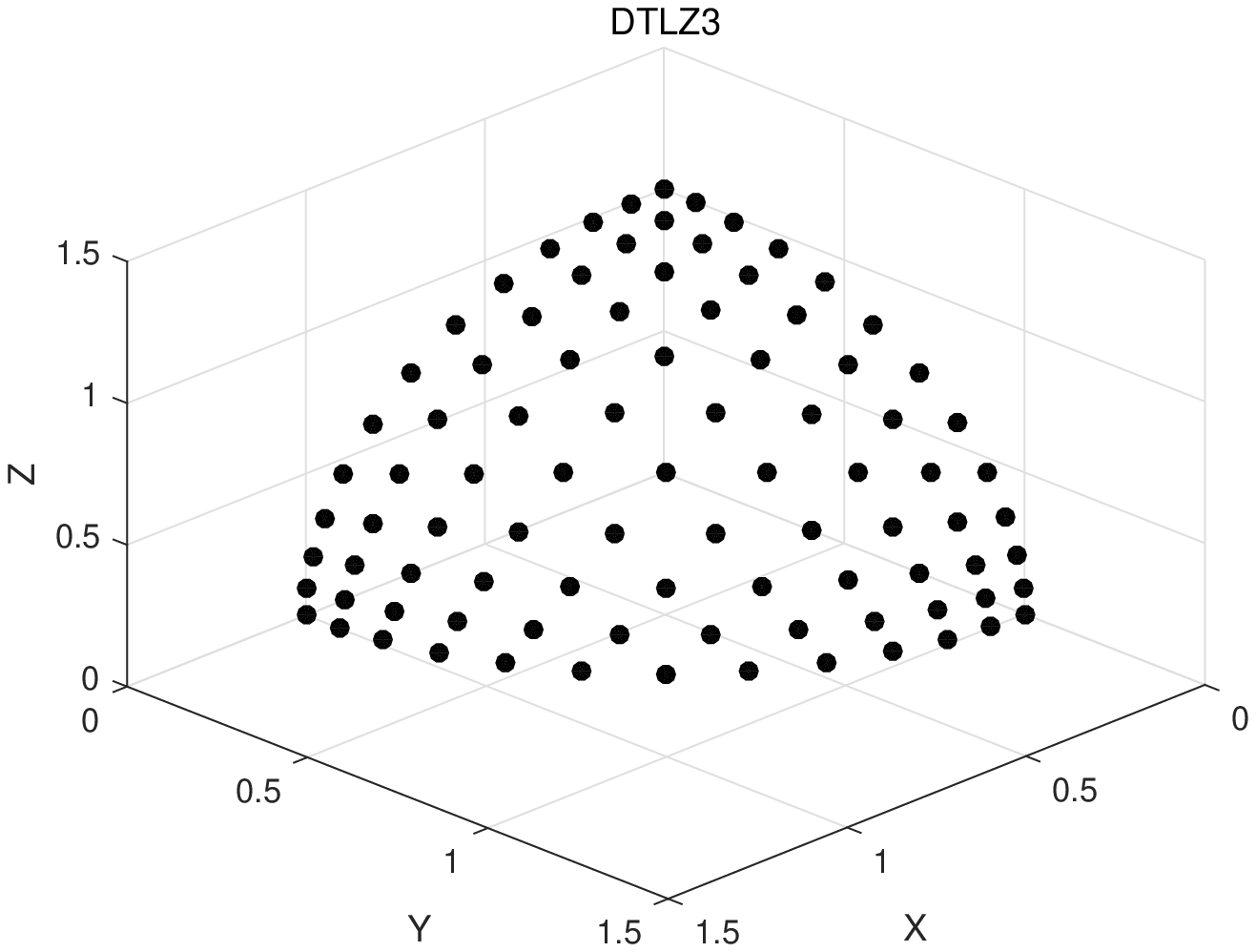}
\end{minipage}
\begin{minipage}[t]{0.5\linewidth}
    \centering
    \includegraphics[width=\textwidth]{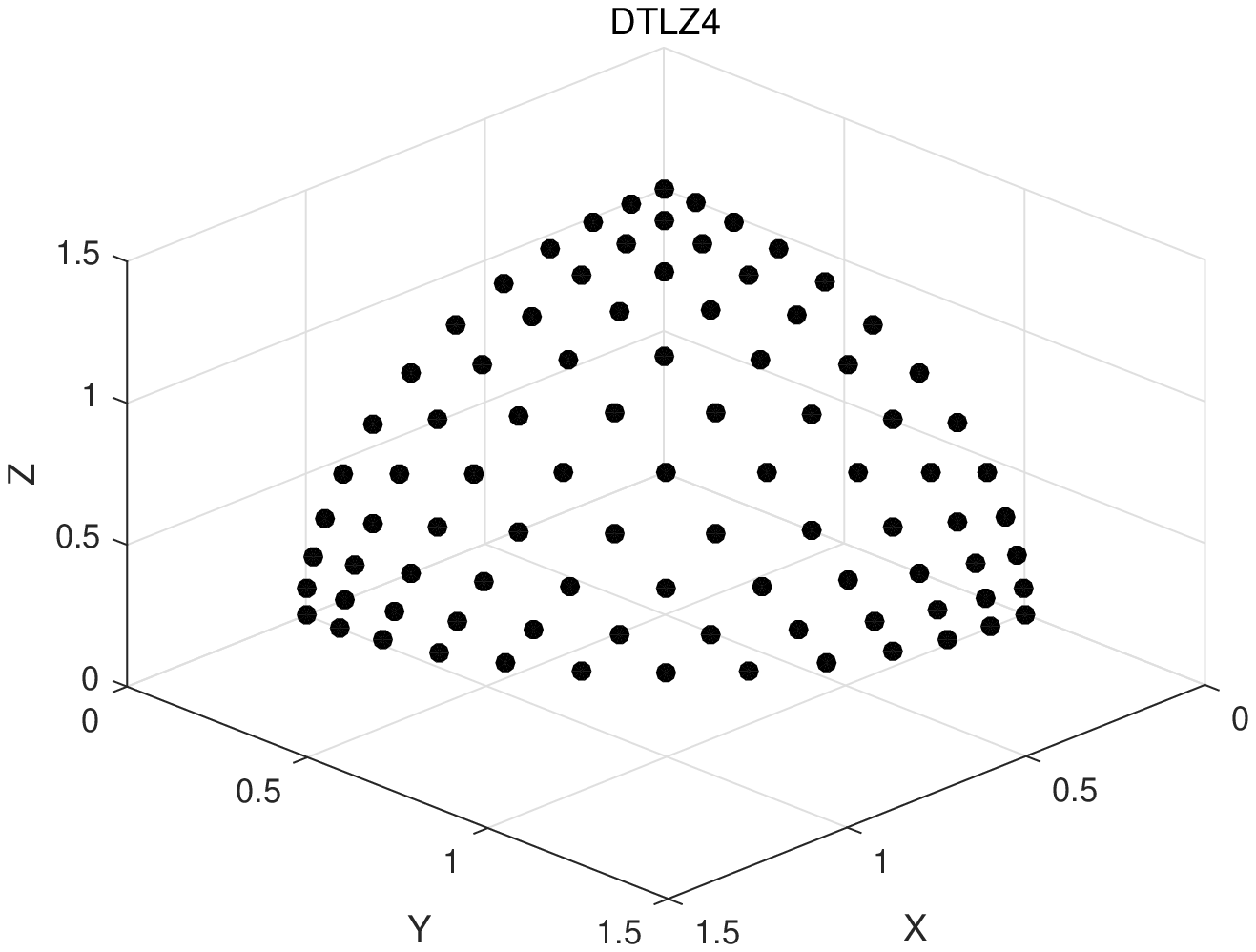}
\end{minipage}
\caption{Non-dominated fronts obtained by MOEA/D-LIU on the 3-objective instance of DTLZ1, DTLZ2, DTLZ3 and DTLZ4, respectively, in the run associated with the median IGD value.}
\label{fig:DTLZM3}
\end{figure*}

\begin{figure*}[!htbp]
\begin{minipage}[t]{0.24\linewidth}
    \centering
    \includegraphics[width=\textwidth]{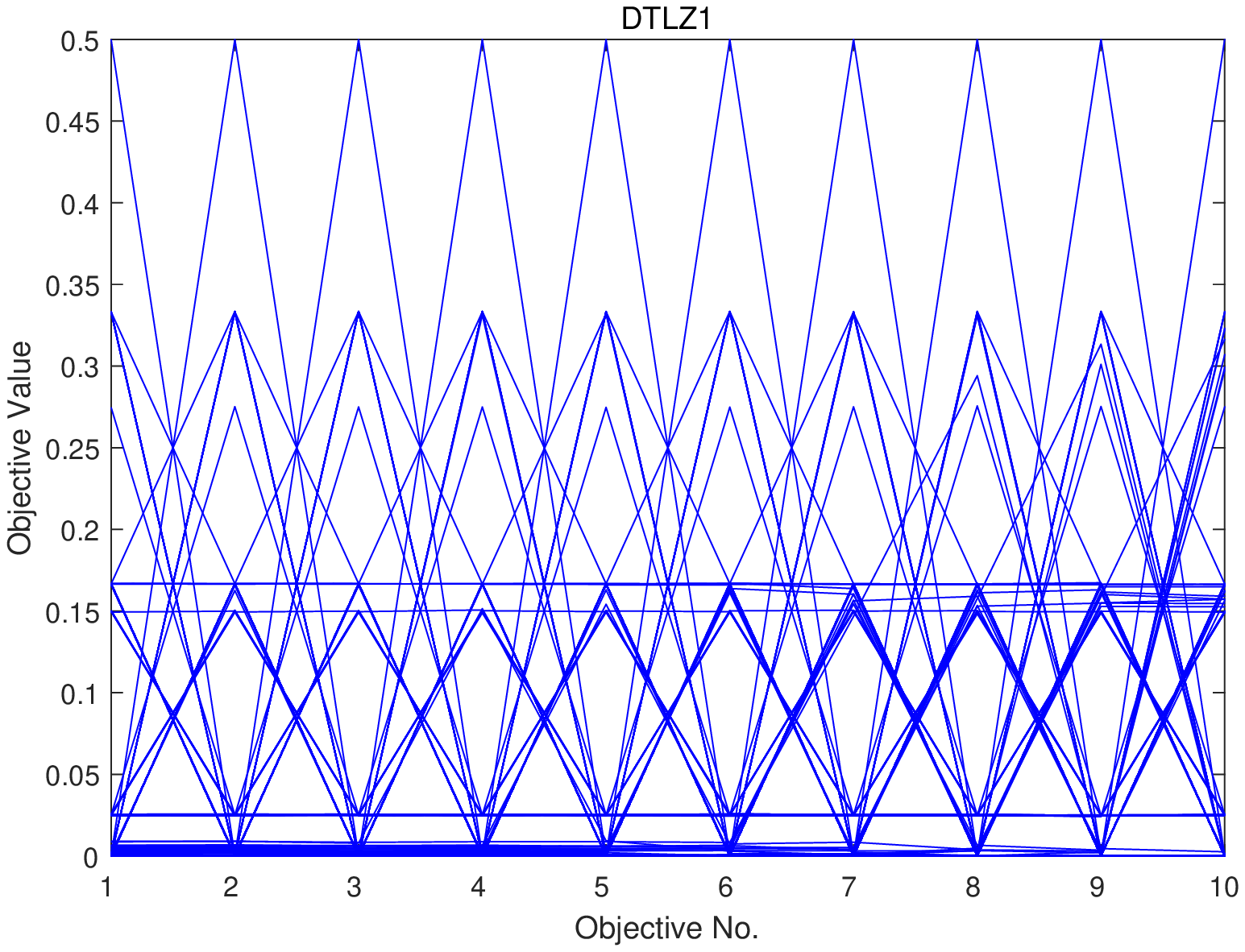}
\end{minipage}
\begin{minipage}[t]{0.24\linewidth}
    \centering
    \includegraphics[width=\textwidth]{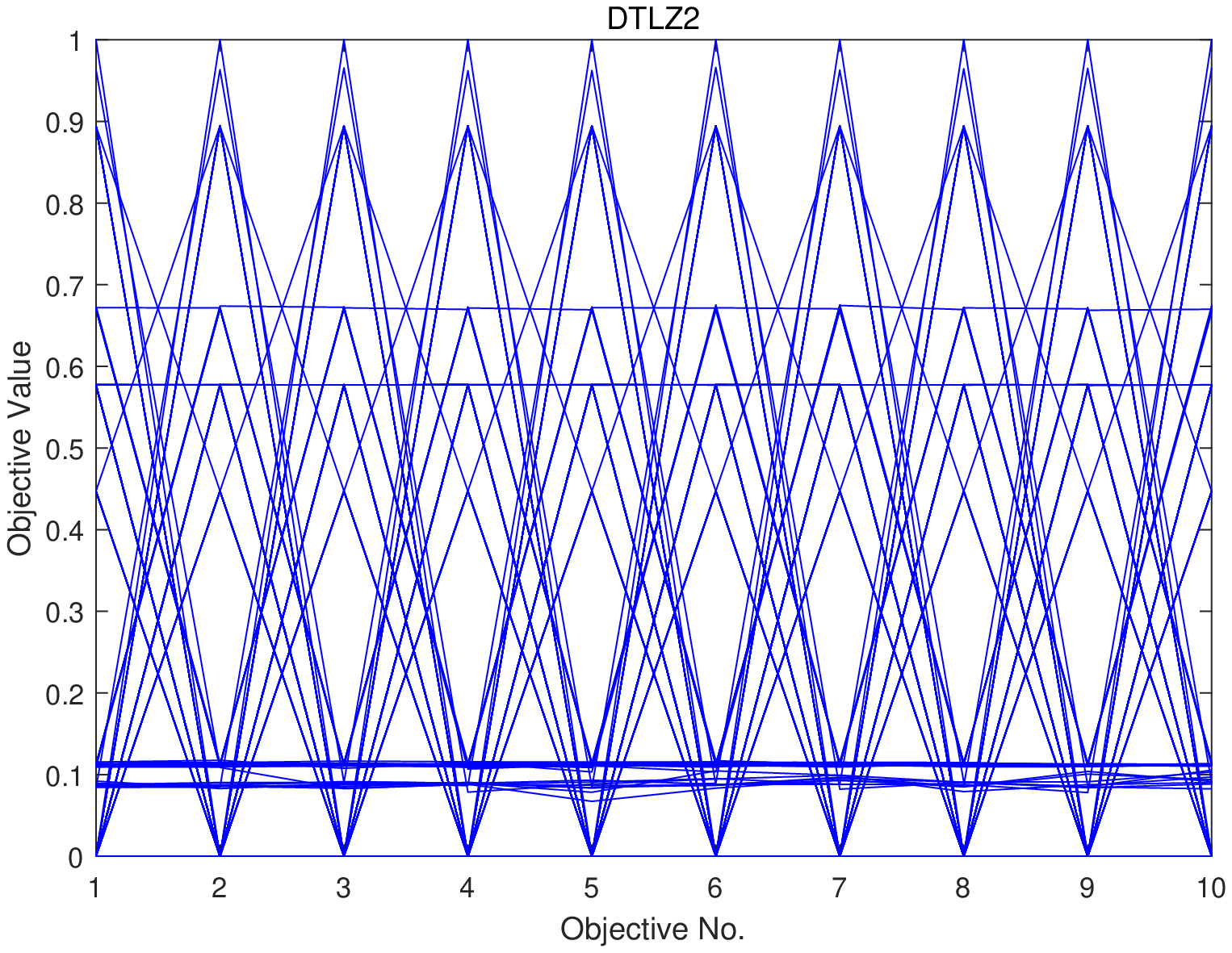}
\end{minipage}
\begin{minipage}[t]{0.24\linewidth}
    \centering
    \includegraphics[width=\textwidth]{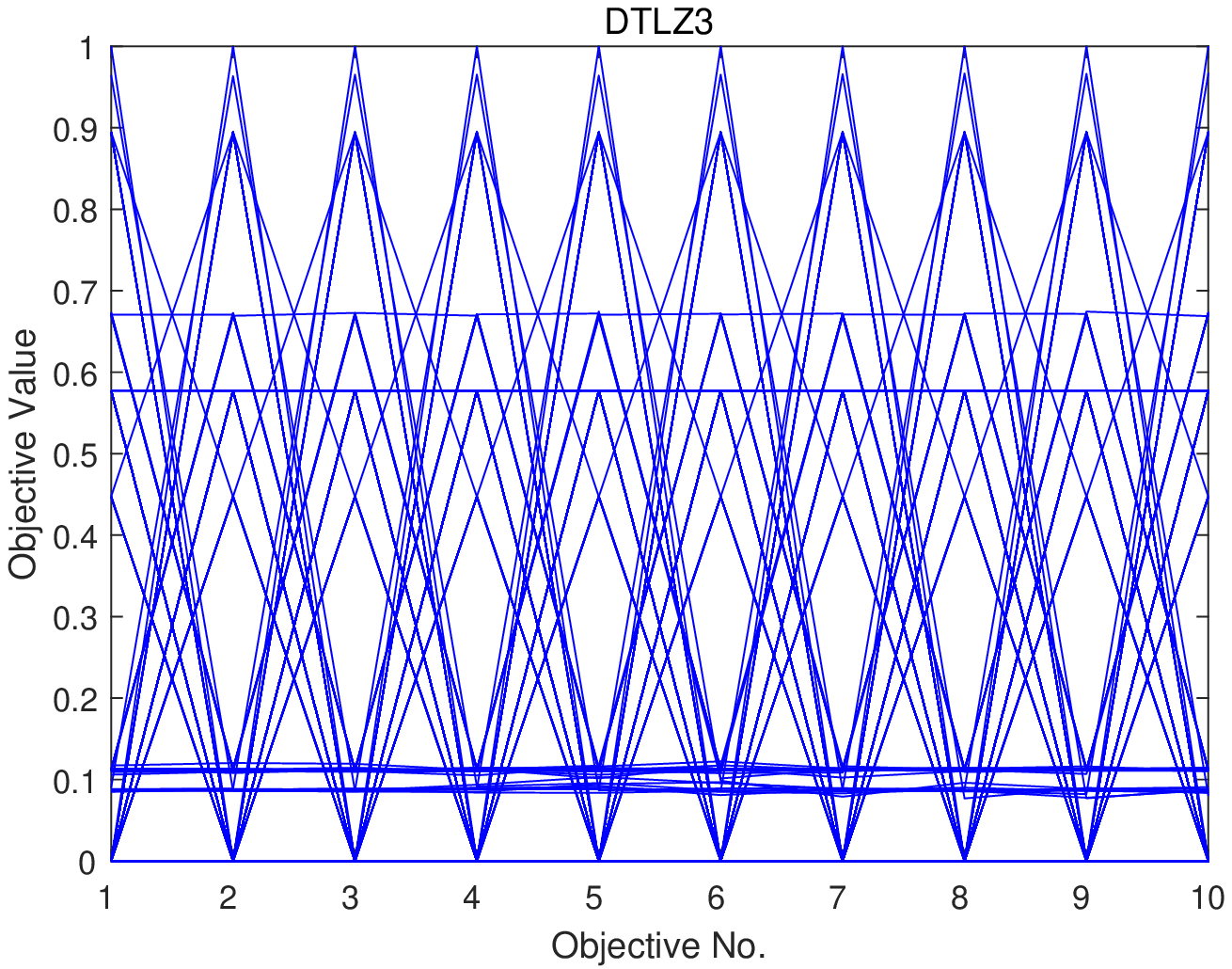}
\end{minipage}
\begin{minipage}[t]{0.24\linewidth}
    \centering
    \includegraphics[width=\textwidth]{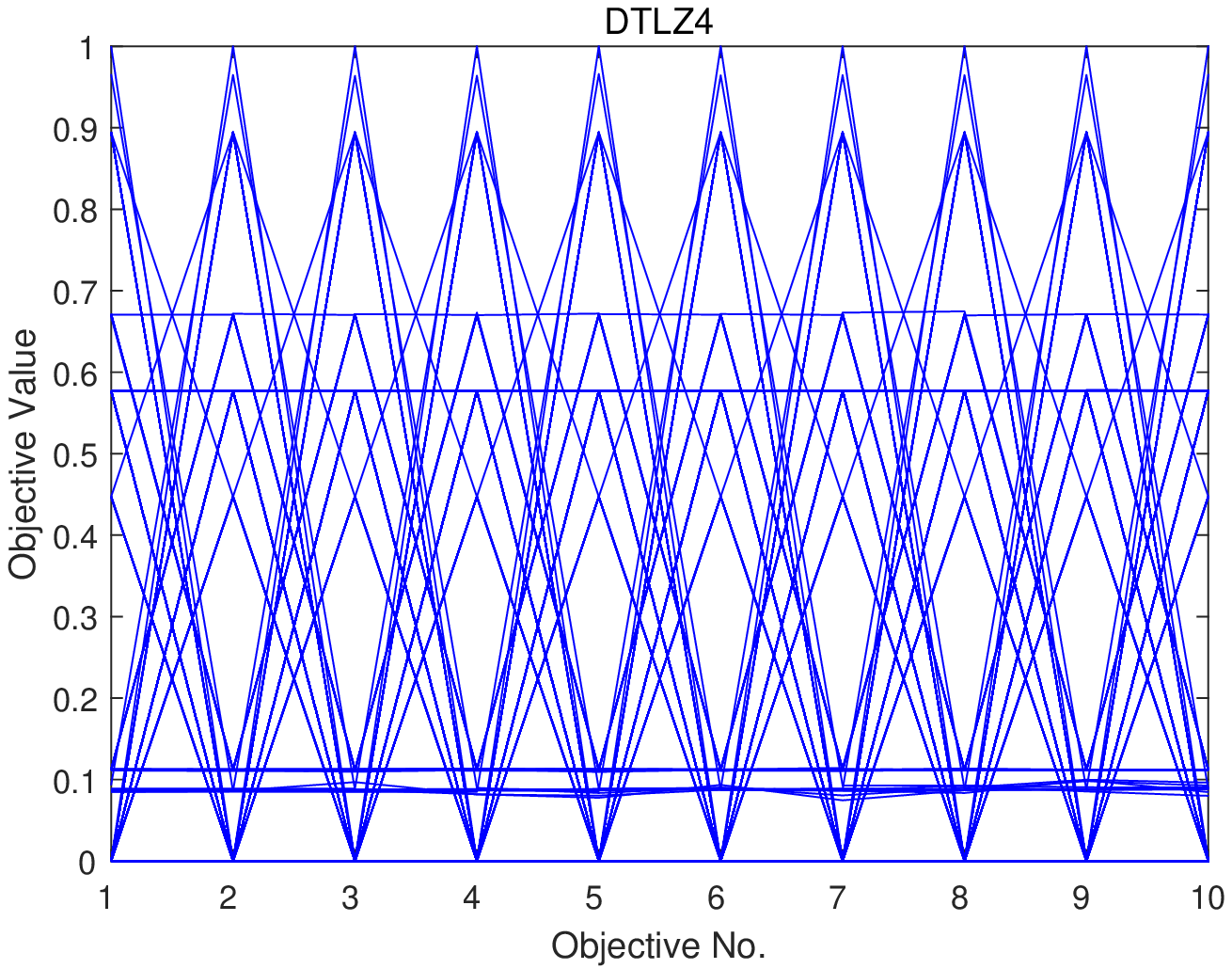}
\end{minipage}
\caption{Parallel coordinates of non-dominated fronts obtained by MOEA/D-LIU on the 10-objective instance of DTLZ1, DTLZ2, DTLZ3 and DTLZ4, respectively, in the run associated with the median IGD value.}
\label{fig:DTLZM10}
\end{figure*}

\begin{figure*}[!htbp]
\begin{minipage}[t]{0.24\linewidth}
    \centering
    \includegraphics[width=\textwidth]{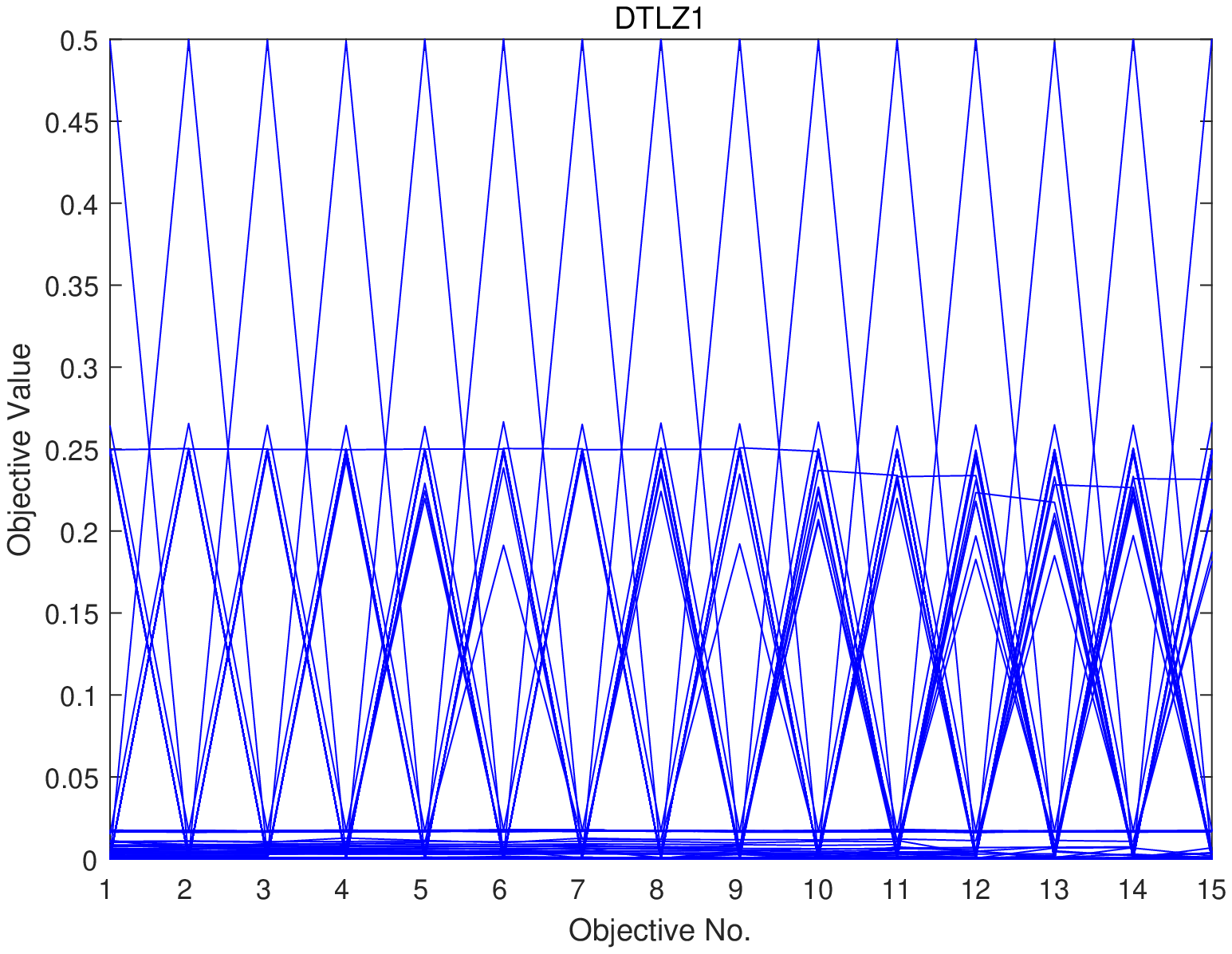}
\end{minipage}
\begin{minipage}[t]{0.24\linewidth}
    \centering
    \includegraphics[width=\textwidth]{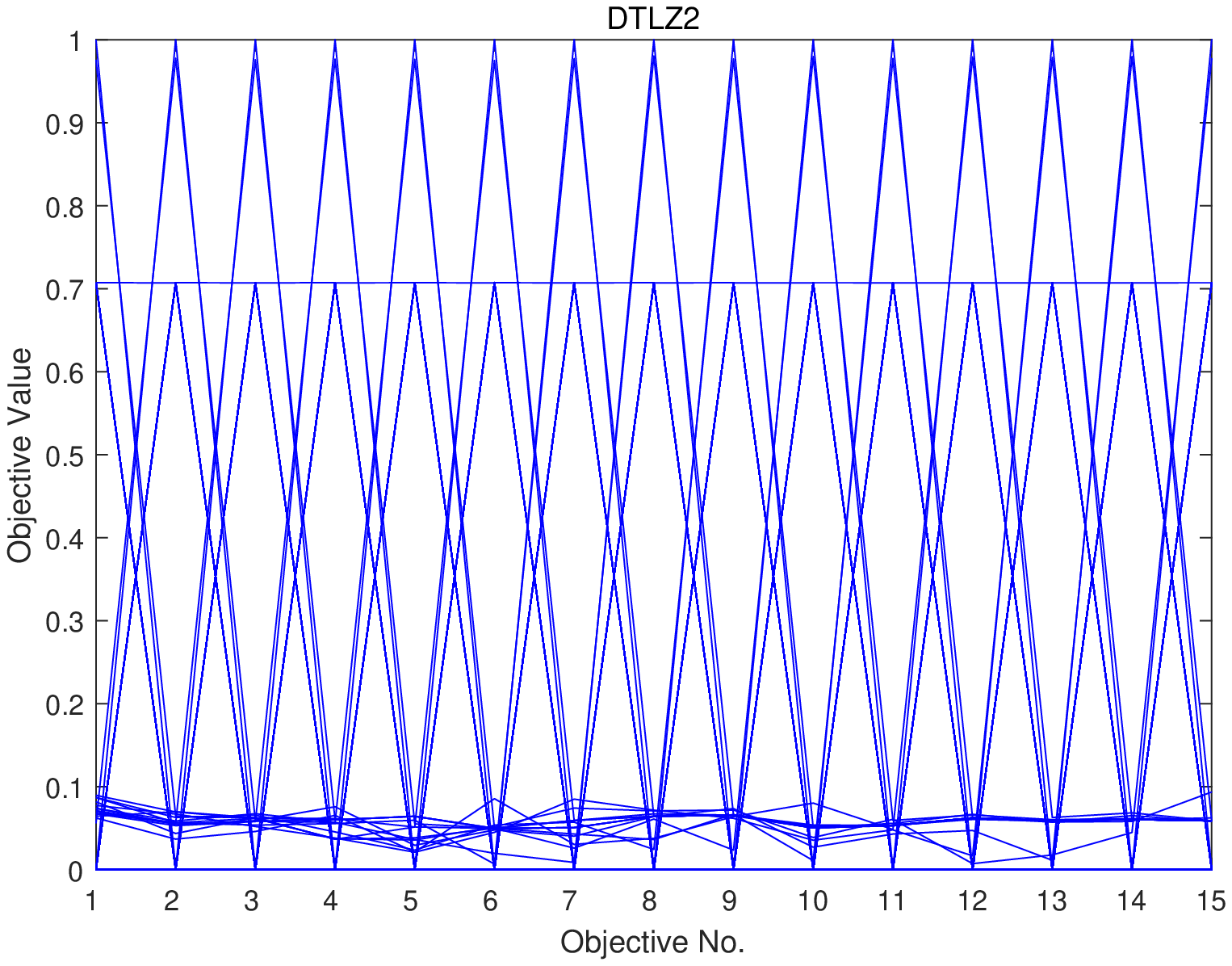}
\end{minipage}
\begin{minipage}[t]{0.24\linewidth}
    \centering
    \includegraphics[width=\textwidth]{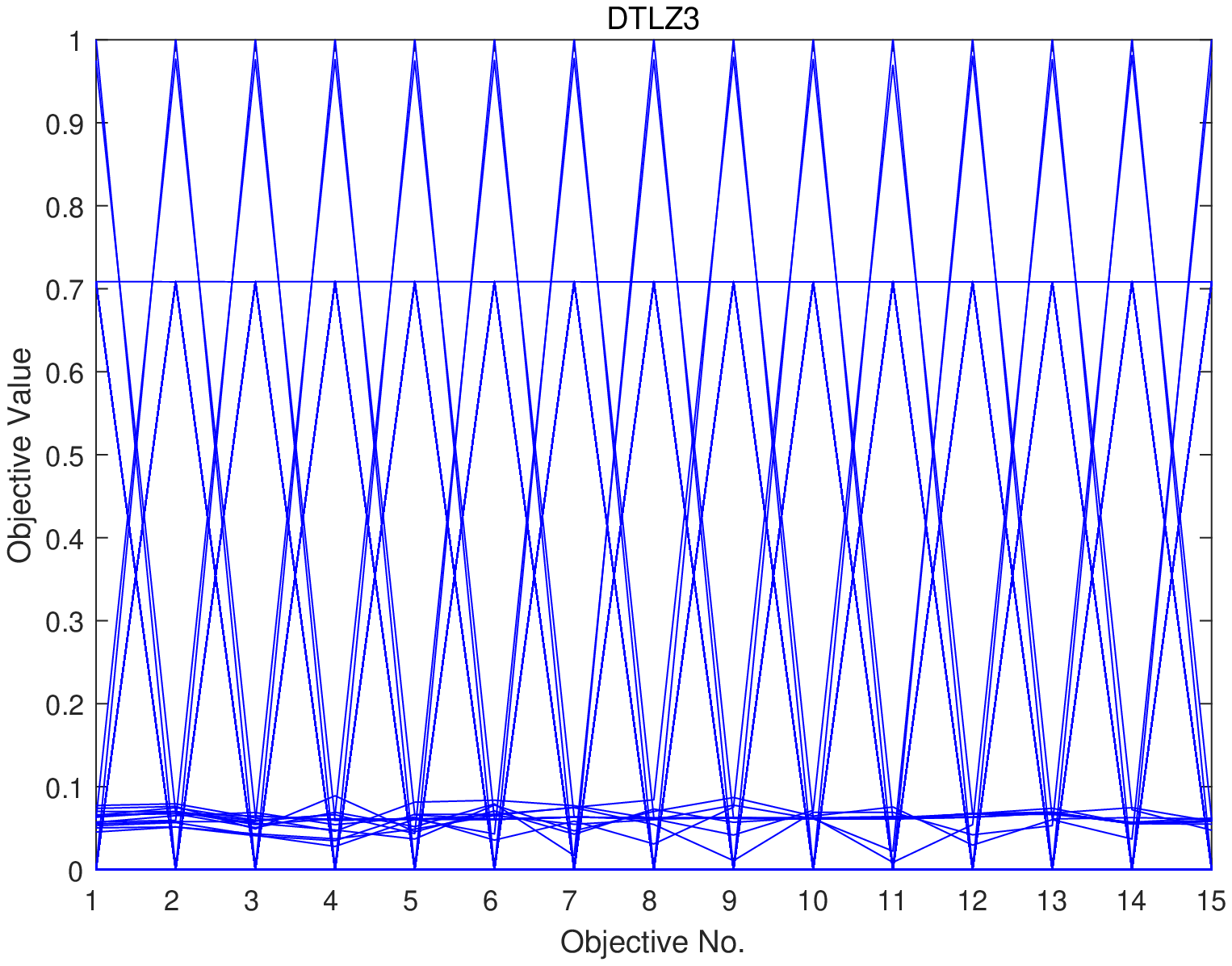}
\end{minipage}
\begin{minipage}[t]{0.24\linewidth}
    \centering
    \includegraphics[width=\textwidth]{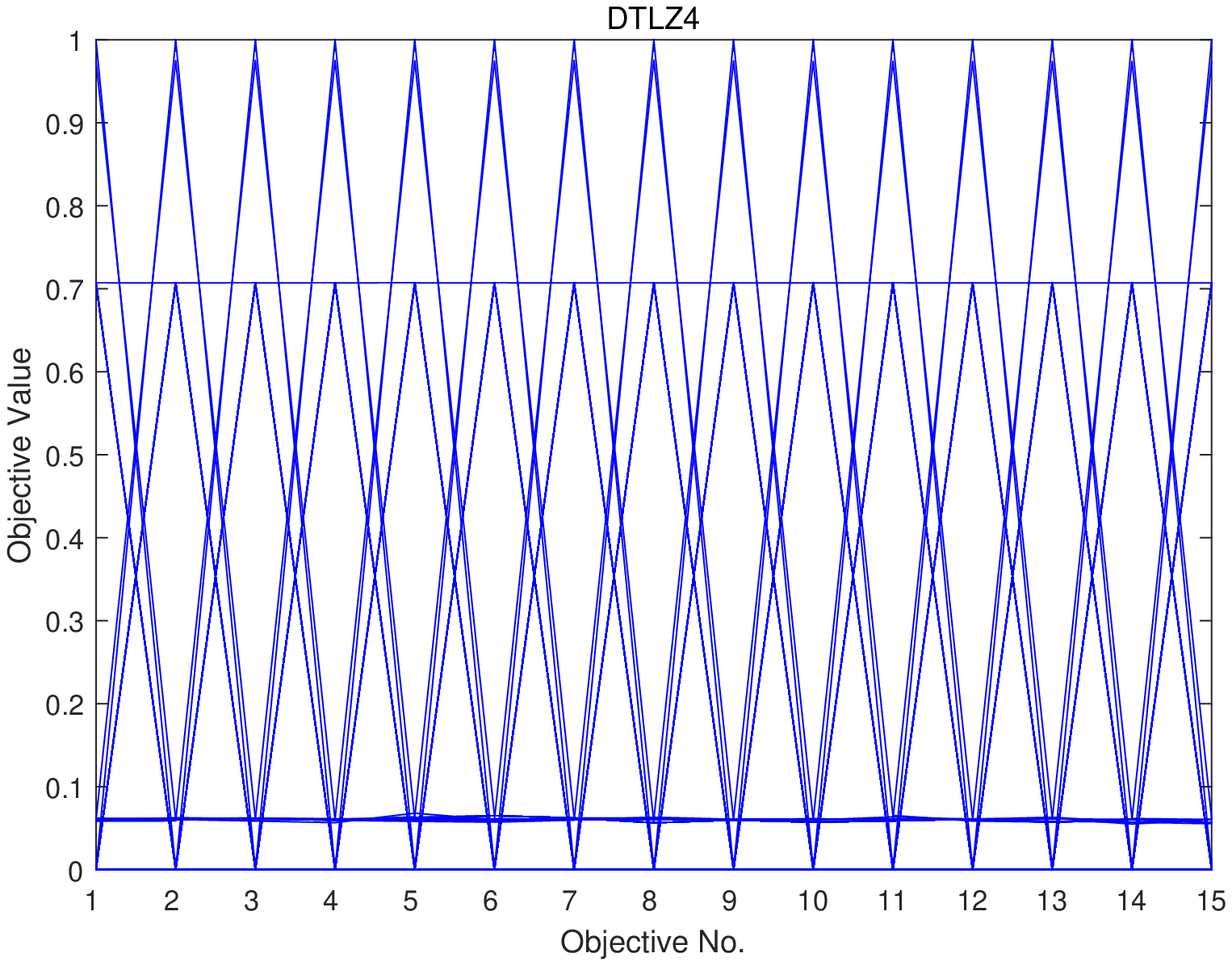}
\end{minipage}
\caption{Parallel coordinates of non-dominated fronts obtained by MOEA/D-LIU on the 15-objective instance of DTLZ1, DTLZ2, DTLZ3 and DTLZ4, respectively, in the run associated with the median IGD value.}
\label{fig:DTLZM15}
\end{figure*}

\begin{table*}[!htbp]
  \centering
  \caption{Best, Median and Worst IGD Values by MOEA/D-LIU, MOEA/DD and MOEA/D
on the instances of DTLZ1, DTLZ2, DTLZ3 and DTLZ4  with Different Number of Objectives.
The values in gray are the best.}
  \resizebox{\linewidth}{!}{
    \begin{tabular}{|c|c|c|c|c|c|c|c|c|}
    \toprule
    Instance & M     & MOEA/D-LIU & MOEA/DD & MOEA/D & Instance & MOEA/D-LIU & MOEA/DD & MOEA/D \\
    \midrule
    \multirow{15}[10]{*}{DTLZ1} & \multirow{3}[2]{*}{3} & \cellcolor[rgb]{ .851,  .851,  .851}2.935E-04 & 3.191E-04 & 4.095E-04 & \multirow{15}[10]{*}{DTLZ2} & \cellcolor[rgb]{ .851,  .851,  .851}3.002E-04 & 6.666E-04 & 5.432E-04 \\
          &       & 1.642E-03 & \cellcolor[rgb]{ .851,  .851,  .851}5.848E-04 & 1.495E-03 &       & \cellcolor[rgb]{ .851,  .851,  .851}3.967E-04 & 8.073E-04 & 6.406E-04 \\
          &       & 5.914E-03 & \cellcolor[rgb]{ .851,  .851,  .851}6.573E-04 & 4.743E-03 &       & \cellcolor[rgb]{ .851,  .851,  .851}5.008E-04 & 1.243E-03 & 8.006E-04 \\
\cmidrule{2-5}\cmidrule{7-9}          & \multirow{3}[2]{*}{5} & \cellcolor[rgb]{ .851,  .851,  .851}2.630E-04 & 2.635E-04 & 3.179E-04 &       & \cellcolor[rgb]{ .851,  .851,  .851}7.014E-04 & 1.128E-03 & 1.219E-03 \\
          &       & 3.316E-04 & \cellcolor[rgb]{ .851,  .851,  .851}2.916E-04 & 6.372E-04 &       & \cellcolor[rgb]{ .851,  .851,  .851}8.221E-04 & 1.291E-03 & 1.437E-03 \\
          &       & 8.378E-04 & \cellcolor[rgb]{ .851,  .851,  .851}3.109E-04 & 1.635E-03 &       & \cellcolor[rgb]{ .851,  .851,  .851}9.059E-04 & 1.424E-03 & 1.727E-03 \\
\cmidrule{2-5}\cmidrule{7-9}          & \multirow{3}[2]{*}{8} & \cellcolor[rgb]{ .851,  .851,  .851}1.508E-03 & 1.809E-03 & 3.914E-03 &       & \cellcolor[rgb]{ .851,  .851,  .851}2.113E-03 & 2.880E-03 & 3.097E-03 \\
          &       & 2.690E-03 & \cellcolor[rgb]{ .851,  .851,  .851}2.589E-03 & 6.106E-03 &       & \cellcolor[rgb]{ .851,  .851,  .851}2.462E-03 & 3.291E-03 & 3.763E-03 \\
          &       & 5.354E-03 & \cellcolor[rgb]{ .851,  .851,  .851}2.996E-03 & 8.537E-03 &       & \cellcolor[rgb]{ .851,  .851,  .851}3.173E-03 & 4.106E-03 & 5.198E-03 \\
\cmidrule{2-5}\cmidrule{7-9}          & \multirow{3}[2]{*}{10} & 1.903E-03 & \cellcolor[rgb]{ .851,  .851,  .851}1.828E-03 & 3.872E-03 &       & \cellcolor[rgb]{ .851,  .851,  .851}1.059E-03 & 3.223E-03 & 2.474E-03 \\
          &       & 3.092E-03 & \cellcolor[rgb]{ .851,  .851,  .851}2.225E-03 & 5.073E-03 &       & \cellcolor[rgb]{ .851,  .851,  .851}1.401E-03 & 3.752E-03 & 2.778E-03 \\
          &       & 4.230E-03 & \cellcolor[rgb]{ .851,  .851,  .851}2.467E-03 & 6.130E-03 &       & \cellcolor[rgb]{ .851,  .851,  .851}1.725E-03 & 4.145E-03 & 3.235E-03 \\
\cmidrule{2-5}\cmidrule{7-9}          & \multirow{3}[2]{*}{15} & 3.661E-03 & \cellcolor[rgb]{ .851,  .851,  .851}2.867E-03 & 1.236E-02 &       & \cellcolor[rgb]{ .851,  .851,  .851}1.320E-03 & 4.557E-03 & 5.254E-03 \\
          &       & 6.573E-03 & \cellcolor[rgb]{ .851,  .851,  .851}4.203E-03 & 1.431E-02 &       & \cellcolor[rgb]{ .851,  .851,  .851}1.810E-03 & 5.863E-03 & 6.005E-03 \\
          &       & 1.000E-02 & \cellcolor[rgb]{ .851,  .851,  .851}4.669E-03 & 1.692E-02 &       & \cellcolor[rgb]{ .851,  .851,  .851}2.488E-03 & 6.929E-03 & 9.409E-03 \\
    \midrule
    \multirow{15}[10]{*}{DTLZ3} & \multirow{3}[2]{*}{} & \cellcolor[rgb]{ .851,  .851,  .851}2.845E-04 & 5.690E-04 & 9.773E-04 & \multirow{15}[10]{*}{DTLZ4} & \cellcolor[rgb]{ .851,  .851,  .851}6.631E-05 & 1.025E-04 & 2.929E-01 \\
          &       & 2.577E-03 & \cellcolor[rgb]{ .851,  .851,  .851}1.892E-03 & 3.426E-03 &       & \cellcolor[rgb]{ .851,  .851,  .851}9.068E-05 & 1.429E-04 & 4.280E-01 \\
          &       & 7.085E-03 & \cellcolor[rgb]{ .851,  .851,  .851}6.231E-03 & 9.113E-03 &       & 5.306E-01 & \cellcolor[rgb]{ .851,  .851,  .851}1.881E-04 & 5.234E-01 \\
\cmidrule{2-5}\cmidrule{7-9}          & \multirow{3}[2]{*}{} & \cellcolor[rgb]{ .851,  .851,  .851}1.500E-04 & 6.181E-04 & 1.129E-03 &       & \cellcolor[rgb]{ .851,  .851,  .851}7.226E-05 & 1.097E-04 & 1.080E-01 \\
          &       & \cellcolor[rgb]{ .851,  .851,  .851}5.190E-04 & 1.181E-03 & 2.213E-03 &       & \cellcolor[rgb]{ .851,  .851,  .851}8.749E-05 & 1.296E-04 & 5.787E-01 \\
          &       & \cellcolor[rgb]{ .851,  .851,  .851}2.765E-03 & 4.736E-03 & 6.147E-03 &       & \cellcolor[rgb]{ .851,  .851,  .851}1.375E-04 & 1.532E-04 & 7.348E-01 \\
\cmidrule{2-5}\cmidrule{7-9}          & \multirow{3}[2]{*}{} & \cellcolor[rgb]{ .851,  .851,  .851}1.906E-03 & 3.411E-03 & 6.459E-03 &       & \cellcolor[rgb]{ .851,  .851,  .851}5.336E-04 & 5.271E-04 & 5.298E-01 \\
          &       & \cellcolor[rgb]{ .851,  .851,  .851}4.378E-03 & 8.079E-03 & 1.948E-02 &       & \cellcolor[rgb]{ .851,  .851,  .851}7.235E-04 & 6.699E-04 & 8.816E-01 \\
          &       & \cellcolor[rgb]{ .851,  .851,  .851}1.455E-02 & 1.826E-02 & 1.123E+00 &       & 9.654E-04 & \cellcolor[rgb]{ .851,  .851,  .851}9.107E-04 & 9.723E-01 \\
\cmidrule{2-5}\cmidrule{7-9}          & \multirow{3}[2]{*}{} & \cellcolor[rgb]{ .851,  .851,  .851}7.356E-04 & 1.689E-03 & 2.791E-03 &       & \cellcolor[rgb]{ .851,  .851,  .851}3.469E-04 & 1.291E-03 & 3.966E-01 \\
          &       & \cellcolor[rgb]{ .851,  .851,  .851}1.004E-03 & 2.164E-03 & 4.319E-03 &       & \cellcolor[rgb]{ .851,  .851,  .851}4.257E-04 & 1.615E-03 & 9.203E-01 \\
          &       & \cellcolor[rgb]{ .851,  .851,  .851}2.089E-03 & 3.226E-03 & 1.010E+00 &       & \cellcolor[rgb]{ .851,  .851,  .851}4.908E-04 & 1.931E-03 & 1.077E+00 \\
\cmidrule{2-5}\cmidrule{7-9}          & \multirow{3}[2]{*}{} & \cellcolor[rgb]{ .851,  .851,  .851}1.392E-03 & 5.716E-03 & 4.360E-03 &       & \cellcolor[rgb]{ .851,  .851,  .851}1.839E-04 & 1.474E-03 & 5.890E-01 \\
          &       & \cellcolor[rgb]{ .851,  .851,  .851}2.163E-03 & 7.461E-03 & 1.664E-02 &       & \cellcolor[rgb]{ .851,  .851,  .851}3.670E-04 & 1.881E-03 & 1.133E+00 \\
          &       & \cellcolor[rgb]{ .851,  .851,  .851}7.695E-03 & 1.138E-02 & 1.260E+00 &       & 1.062E-01 & \cellcolor[rgb]{ .851,  .851,  .851}3.159E-03 & 1.249E+00 \\
    \bottomrule
    \end{tabular}}%
  \label{IGDonDTLZ}%
\end{table*}%



All instances  run 20 times independently and their average running times are listed in Table \ref{tab:AVGRuntimesOfDTLZ}.
As it can be seen, the average running times of MOEA/D-LIU on all instances are less than those of MOEA/D and MOEA/DD, except those of the 15-objective instances of DTLZ3 and DTLZ4.
This indicates that MOEA/D-LIU is computationally efficient on these problems.

The non-dominated fronts, obtained by MOEA/D-LIU on the 3-objective instances of the four problems in the run associated with the median IGD value, are plotted in Fig. \ref{fig:DTLZM3}.
As can be seen from the figure, MOEA/D-LIU performs well on all the 3-objective instances, in terms of convergence and distribution.

Fig. \ref{fig:DTLZM10} and \ref{fig:DTLZM15} show the parallel coordinates of non-dominated fronts obtained by MOEA/D-LIU, on the 10- and 15-objective instances of DTLZ1 to DTLZ4, respectively, in the run associated with median IGD value. It can be observed that the non-dominated fronts obtained by the algorithm are promising in both convergence and distribution. As it is mentioned before, the four problems have many local minima in the search space to hinder a MOEA to converge to the global PF.  Therefore, our observations mean that the algorithm has good performance to cross the local minima and approach the PF.

We calculate the IGD values of the solution sets found by MOEA/D-LIU,
and compare the calculation results with those presented in \citep{MOEADD}, where the experimental results obtained by MOEA/DD, MOEA/D, NSGA-III, GrEA and HyPE are compared.
But only the experimental results of MOEA/DD and MOEA/D are preserved for comparison, since NSGA-III, GrEA and HyPE do not win on all of the comparisons in that paper.
The detailed data are shown in Table \ref{IGDonDTLZ}.
\begin{itemize}
  \item DTLZ1:It can be seen from  Table \ref{IGDonDTLZ} that MOEA/D-LIU and MOEA/DD perform better than MOEA/D on all of the IGD values. In detail, MOEA/DD wins 12 out of 15 values and wins all comparisons of the 10- and 15-objective instances, and can be considered as the best optimizer for DTLZ1.
  \item DTLZ2:The performance of MOEA/D-LIU, MOEA/DD and MOEA/D is comparable in this problem. From the specific data, MOEA/D-LIU shows the best performance on all of the instances and is the best optimizer for DTLZ2.
  \item DTLZ3:Although MOEA/DD performs well on all of the instances of DTLZ3, MOEA/D-LIU wins 13 out of 15 comparisons, and is the best optimizer for this problem. It seems that MOEA/D is not stable on 8-, 10- and 15-objective  instances of this problem, since the difference between the best, median and worst values are significant.
  \item DTLZ4:MOEA/D performs far worse than MOEA/D-LIU and MOEA/DD on this problem, because of the degeneration of its population on all the instances of DTLZ4. To be specific, MOEA/D-LIU wins 12 out 15 comparisons but the population degenerates in several running instances.
\end{itemize}

On the whole, our algorithm can obtain well converged and distributed results on  the four problems.
It wins on 43 out of 60 comparisons, and shows a promising performance on these problems, especially on DTLZ2 to DTLZ4.

\subsection{Performance Comparisons on WFG1 to WFG9}

\begin{table*}[!htbp]
  \centering
  \caption{Running times  (in milliseconds)  of MOEA/D-LIU, MOEA/D and MOEA/DD on the instances of problems WFG1 to WFG9.}
  \resizebox{\linewidth}{!}{
    \begin{tabular}{|c|c|c|c|c|c|c|c|c|c|c|c|c|}
    \toprule
    Problem & M     & MOEA/D-LIU & MOEA/D & MOEA/DD & Problem & MOEA/D-LIU & MOEA/D & MOEA/DD & Problem & MOEA/D-LIU & MOEA/D & MOEA/DD \\
    \midrule
    \multirow{4}[8]{*}{WFG1} & 3     & \cellcolor[rgb]{ .851,  .851,  .851}11101  & 13688  & 12998  & \multirow{4}[8]{*}{WFG2} & \cellcolor[rgb]{ .851,  .851,  .851}11006  & 14974  & 10869  & \multirow{4}[8]{*}{WFG3} & \cellcolor[rgb]{ .851,  .851,  .851}10288  & 11613  & 12114  \\
\cmidrule{2-5}\cmidrule{7-9}\cmidrule{11-13}          & 5     & \cellcolor[rgb]{ .851,  .851,  .851}27909  & 31442  & 72972  &       & \cellcolor[rgb]{ .851,  .851,  .851}29996  & 36821  & 63353  &       & \cellcolor[rgb]{ .851,  .851,  .851}28760  & 34948  & 74802  \\
\cmidrule{2-5}\cmidrule{7-9}\cmidrule{11-13}          & 8     & \cellcolor[rgb]{ .851,  .851,  .851}29535  & 36671  & 57140  &       & \cellcolor[rgb]{ .851,  .851,  .851}26808  & 32071  & 48192  &       & \cellcolor[rgb]{ .851,  .851,  .851}25678  & 31521  & 51899  \\
\cmidrule{2-5}\cmidrule{7-9}\cmidrule{11-13}          & 10    & \cellcolor[rgb]{ .851,  .851,  .851}57962  & 70712  & 189756  &       & \cellcolor[rgb]{ .851,  .851,  .851}56319  & 71433  & 167618  &       & \cellcolor[rgb]{ .851,  .851,  .851}51985  & 69271  & 179002  \\
    \midrule
    \multirow{4}[8]{*}{WFG4} & 3     & \cellcolor[rgb]{ .851,  .851,  .851}11531  & 12392  & 11818  & \multirow{4}[8]{*}{WFG5} & \cellcolor[rgb]{ .851,  .851,  .851}10882  & 12175  & 9443  & \multirow{4}[8]{*}{WFG6} & \cellcolor[rgb]{ .851,  .851,  .851}10991  & 12554  & 9005  \\
\cmidrule{2-5}\cmidrule{7-9}\cmidrule{11-13}          & 5     & \cellcolor[rgb]{ .851,  .851,  .851}29966  & 36898  & 65820  &       & \cellcolor[rgb]{ .851,  .851,  .851}29749  & 36936  & 60567  &       & \cellcolor[rgb]{ .851,  .851,  .851}30194  & 38282  & 61368  \\
\cmidrule{2-5}\cmidrule{7-9}\cmidrule{11-13}          & 8     & \cellcolor[rgb]{ .851,  .851,  .851}25172  & 27825  & 49954  &       & \cellcolor[rgb]{ .851,  .851,  .851}25477  & 32039  & 46549  &       & \cellcolor[rgb]{ .851,  .851,  .851}29329  & 37108  & 46222  \\
\cmidrule{2-5}\cmidrule{7-9}\cmidrule{11-13}          & 10    & \cellcolor[rgb]{ .851,  .851,  .851}59459  & 74637  & 172702  &       & \cellcolor[rgb]{ .851,  .851,  .851}43072  & 55055  & 159943  &       & \cellcolor[rgb]{ .851,  .851,  .851}58963  & 78503  & 160831  \\
    \midrule
    \multirow{4}[8]{*}{WFG7} & 3     & \cellcolor[rgb]{ .851,  .851,  .851}11081  & 12413  & 10460  & \multirow{4}[8]{*}{WFG8} & 9406  & \cellcolor[rgb]{ .851,  .851,  .851}9290  & 12745  & \multirow{4}[8]{*}{WFG9} & \cellcolor[rgb]{ .851,  .851,  .851}13811  & 16496  & 13181  \\
\cmidrule{2-5}\cmidrule{7-9}\cmidrule{11-13}          & 5     & \cellcolor[rgb]{ .851,  .851,  .851}32461  & 41428  & 65038  &       & \cellcolor[rgb]{ .851,  .851,  .851}33923  & 41670  & 65441  &       & \cellcolor[rgb]{ .851,  .851,  .851}35070  & 44804  & 69845  \\
\cmidrule{2-5}\cmidrule{7-9}\cmidrule{11-13}          & 8     & \cellcolor[rgb]{ .851,  .851,  .851}29792  & 34446  & 48423  &       & \cellcolor[rgb]{ .851,  .851,  .851}30154  & 33167  & 47910  &       & \cellcolor[rgb]{ .851,  .851,  .851}34728  & 42858  & 56149  \\
\cmidrule{2-5}\cmidrule{7-9}\cmidrule{11-13}          & 10    & \cellcolor[rgb]{ .851,  .851,  .851}55803  & 65925  & 153036  &       & \cellcolor[rgb]{ .851,  .851,  .851}72728  & 83338  & 163188  &       & \cellcolor[rgb]{ .851,  .851,  .851}65139  & 84386  & 184501  \\
    \bottomrule
    \end{tabular}}%
  \label{tab:AVGRuntimesOnWFG}%
\end{table*}%

\begin{table*}[!htbp]
  \centering
  \caption{Best, median and worst HV values by MOEA/D-LIU, MOEA/DD and GrEA on the instances of problems WFG1 to WFG9 with different number of objectives. The values in gray are the best.}
    \resizebox{\linewidth}{!}{
    \begin{tabular}{|c|c|c|c|c|c|c|c|c|c|c|c|c|}
    \toprule
    Instance & M     & MOEA/D-LIU & MOEA/DD & GrEA  & Instance & MOEA/D-LIU & MOEA/DD & GrEA  & Instance & MOEA/D-LIU & MOEA/DD & GrEA \\
    \midrule
    \multirow{12}[8]{*}{WFG1} & \multirow{3}[2]{*}{3} & 0.934327  & \cellcolor[rgb]{ .851,  .851,  .851}0.937694  & 0.794748  & \multirow{12}[8]{*}{WFG2} & 0.955623  & \cellcolor[rgb]{ .851,  .851,  .851}0.958287  & 0.950084  & \multirow{12}[8]{*}{WFG3} & \cellcolor[rgb]{ .851,  .851,  .851}0.712329  & 0.703664  & 0.699502  \\
          &       & 0.927851  & \cellcolor[rgb]{ .851,  .851,  .851}0.933402  & 0.692567  &       & 0.945868  & \cellcolor[rgb]{ .851,  .851,  .851}0.952467  & 0.942908  &       & \cellcolor[rgb]{ .851,  .851,  .851}0.708537  & 0.702964  & 0.672221  \\
          &       & \cellcolor[rgb]{ .851,  .851,  .851}0.911181  & 0.899253  & 0.627963  &       & \cellcolor[rgb]{ .851,  .851,  .851}0.803567  & 0.803397  & 0.800186  &       & \cellcolor[rgb]{ .851,  .851,  .851}0.702307  & 0.701624  & 0.662046  \\
\cmidrule{2-5}\cmidrule{7-9}\cmidrule{11-13}          & \multirow{3}[2]{*}{5} & 0.894600  & \cellcolor[rgb]{ .851,  .851,  .851}0.963464  & 0.876644  &       & \cellcolor[rgb]{ .851,  .851,  .851}0.995723  & 0.986572  & 0.980806  &       & \cellcolor[rgb]{ .851,  .851,  .851}0.694272  & 0.673031  & 0.695221  \\
          &       & 0.872031  & \cellcolor[rgb]{ .851,  .851,  .851}0.960897  & 0.831814  &       & \cellcolor[rgb]{ .851,  .851,  .851}0.991998  & 0.985129  & 0.976837  &       & \cellcolor[rgb]{ .851,  .851,  .851}0.692238  & 0.668938  & 0.684583  \\
          &       & 0.837747  & \cellcolor[rgb]{ .851,  .851,  .851}0.959840  & 0.790367  &       & 0.813873  & \cellcolor[rgb]{ .851,  .851,  .851}0.980035  & 0.808125  &       & \cellcolor[rgb]{ .851,  .851,  .851}0.690319  & 0.662951  & 0.671553  \\
\cmidrule{2-5}\cmidrule{7-9}\cmidrule{11-13}          & \multirow{3}[2]{*}{8} & 0.842511  & \cellcolor[rgb]{ .851,  .851,  .851}0.922284  & 0.811760  &       & 0.970465  & \cellcolor[rgb]{ .851,  .851,  .851}0.981673  & 0.980012  &       & 0.551058  & 0.598892  & \cellcolor[rgb]{ .851,  .851,  .851}0.657744  \\
          &       & 0.813466  & \cellcolor[rgb]{ .851,  .851,  .851}0.913024  & 0.681959  &       & 0.780563  & \cellcolor[rgb]{ .851,  .851,  .851}0.967265  & 0.840293  &       & 0.532740  & 0.565609  & \cellcolor[rgb]{ .851,  .851,  .851}0.649020  \\
          &       & 0.727155  & \cellcolor[rgb]{ .851,  .851,  .851}0.877784  & 0.616006  &       & 0.765842  & \cellcolor[rgb]{ .851,  .851,  .851}0.789739  & 0.778291  &       & 0.518037  & 0.556725  & \cellcolor[rgb]{ .851,  .851,  .851}0.638147  \\
\cmidrule{2-5}\cmidrule{7-9}\cmidrule{11-13}          & \multirow{3}[2]{*}{10} & 0.879136  & \cellcolor[rgb]{ .851,  .851,  .851}0.926815  & 0.866298  &       & \cellcolor[rgb]{ .851,  .851,  .851}0.968422  & 0.968201  & 0.964235  &       & 0.541417  & \cellcolor[rgb]{ .851,  .851,  .851}0.552713  & 0.543352  \\
          &       & 0.845289  & \cellcolor[rgb]{ .851,  .851,  .851}0.919789  & 0.832016  &       & 0.779356  & \cellcolor[rgb]{ .851,  .851,  .851}0.965345  & 0.959740  &       & 0.513744  & \cellcolor[rgb]{ .851,  .851,  .851}0.532897  & 0.513261  \\
          &       & 0.826000  & \cellcolor[rgb]{ .851,  .851,  .851}0.864689  & 0.757841  &       & 0.769651  & \cellcolor[rgb]{ .851,  .851,  .851}0.961400  & 0.956533  &       & 0.502054  & \cellcolor[rgb]{ .851,  .851,  .851}0.504943  & 0.501210  \\
    \midrule
    \multirow{12}[8]{*}{WFG4} & \multirow{3}[2]{*}{3} & \cellcolor[rgb]{ .851,  .851,  .851}0.728787  & 0.727060  & 0.723403  & \multirow{12}[8]{*}{WFG5} & \cellcolor[rgb]{ .851,  .851,  .851}0.697678  & 0.693665  & 0.689784  & \multirow{12}[8]{*}{WFG6} & \cellcolor[rgb]{ .851,  .851,  .851}0.709895  & 0.708910  & 0.699876  \\
          &       & \cellcolor[rgb]{ .851,  .851,  .851}0.727206  & 0.726927  & 0.722997  &       & \cellcolor[rgb]{ .851,  .851,  .851}0.694220  & 0.693544  & 0.689177  &       & \cellcolor[rgb]{ .851,  .851,  .851}0.700159  & 0.699663  & 0.693984  \\
          &       & 0.726563  & \cellcolor[rgb]{ .851,  .851,  .851}0.726700  & 0.722629  &       & 0.686005  & \cellcolor[rgb]{ .851,  .851,  .851}0.691173  & 0.688885  &       & \cellcolor[rgb]{ .851,  .851,  .851}0.692365  & 0.689125  & 0.685599  \\
\cmidrule{2-5}\cmidrule{7-9}\cmidrule{11-13}          & \multirow{3}[2]{*}{5} & \cellcolor[rgb]{ .851,  .851,  .851}0.885161  & 0.876181  & 0.881161  &       & \cellcolor[rgb]{ .851,  .851,  .851}0.844540  & 0.833159  & 0.836232  &       & \cellcolor[rgb]{ .851,  .851,  .851}0.859830  & 0.850531  & 0.855839  \\
          &       & \cellcolor[rgb]{ .851,  .851,  .851}0.883841  & 0.875836  & 0.879484  &       & \cellcolor[rgb]{ .851,  .851,  .851}0.841753  & 0.832710  & 0.834726  &       & \cellcolor[rgb]{ .851,  .851,  .851}0.850898  & 0.838329  & 0.847137  \\
          &       & \cellcolor[rgb]{ .851,  .851,  .851}0.881809  & 0.875517  & 0.877642  &       & \cellcolor[rgb]{ .851,  .851,  .851}0.837329  & 0.830367  & 0.832212  &       & \cellcolor[rgb]{ .851,  .851,  .851}0.845577  & 0.828315  & 0.840637  \\
\cmidrule{2-5}\cmidrule{7-9}\cmidrule{11-13}          & \multirow{3}[2]{*}{8} & \cellcolor[rgb]{ .851,  .851,  .851}0.945976  & 0.920869  & 0.787287  &       & \cellcolor[rgb]{ .851,  .851,  .851}0.896624  & 0.852838  & 0.838183  &       & \cellcolor[rgb]{ .851,  .851,  .851}0.914666  & 0.876310  & 0.912095  \\
          &       & \cellcolor[rgb]{ .851,  .851,  .851}0.944500  & 0.910146  & 0.784141  &       & \cellcolor[rgb]{ .851,  .851,  .851}0.894250  & 0.846736  & 0.641973  &       & \cellcolor[rgb]{ .851,  .851,  .851}0.902640  & 0.863087  & 0.902638  \\
          &       & \cellcolor[rgb]{ .851,  .851,  .851}0.936960  & 0.902710  & 0.679178  &       & \cellcolor[rgb]{ .851,  .851,  .851}0.887472  & 0.830338  & 0.571933  &       & \cellcolor[rgb]{ .851,  .851,  .851}0.892441  & 0.844535  & 0.885712  \\
\cmidrule{2-5}\cmidrule{7-9}\cmidrule{11-13}          & \multirow{3}[2]{*}{10} & \cellcolor[rgb]{ .851,  .851,  .851}0.974033  & 0.913018  & 0.896261  &       & \cellcolor[rgb]{ .851,  .851,  .851}0.919431  & 0.848321  & 0.791725  &       & 0.939370  & 0.884394  & \cellcolor[rgb]{ .851,  .851,  .851}0.943454  \\
          &       & \cellcolor[rgb]{ .851,  .851,  .851}0.972429  & 0.907040  & 0.843257  &       & \cellcolor[rgb]{ .851,  .851,  .851}0.917859  & 0.841118  & 0.725198  &       & \cellcolor[rgb]{ .851,  .851,  .851}0.928172  & 0.859986  & 0.927443  \\
          &       & \cellcolor[rgb]{ .851,  .851,  .851}0.967384  & 0.888885  & 0.840257  &       & \cellcolor[rgb]{ .851,  .851,  .851}0.910986  & 0.829547  & 0.685882  &       & \cellcolor[rgb]{ .851,  .851,  .851}0.916970  & 0.832299  & 0.884145  \\
    \midrule
    \multirow{12}[8]{*}{WFG7} & \multirow{3}[2]{*}{3} & \cellcolor[rgb]{ .851,  .851,  .851}0.730098  & 0.727069  & 0.723229  & \multirow{12}[8]{*}{WFG8} & \cellcolor[rgb]{ .851,  .851,  .851}0.674122  & 0.672022  & 0.671845  & \multirow{12}[8]{*}{WFG9} & 0.690128  & \cellcolor[rgb]{ .851,  .851,  .851}0.707269  & 0.702489  \\
          &       & \cellcolor[rgb]{ .851,  .851,  .851}0.728677  & 0.727012  & 0.722843  &       & \cellcolor[rgb]{ .851,  .851,  .851}0.671440  & 0.670558  & 0.669762  &       & 0.650441  & \cellcolor[rgb]{ .851,  .851,  .851}0.687401  & 0.638103  \\
          &       & \cellcolor[rgb]{ .851,  .851,  .851}0.727700  & 0.726907  & 0.722524  &       & \cellcolor[rgb]{ .851,  .851,  .851}0.668910  & 0.668593  & 0.667948  &       & \cellcolor[rgb]{ .851,  .851,  .851}0.638940  & 0.638194  & 0.636575  \\
\cmidrule{2-5}\cmidrule{7-9}\cmidrule{11-13}          & \multirow{3}[2]{*}{5} & \cellcolor[rgb]{ .851,  .851,  .851}0.888179  & 0.876409  & 0.884174  &       & \cellcolor[rgb]{ .851,  .851,  .851}0.817293  & 0.818663  & 0.797496  &       & 0.819551  & \cellcolor[rgb]{ .851,  .851,  .851}0.834616  & 0.823916  \\
          &       & \cellcolor[rgb]{ .851,  .851,  .851}0.887903  & 0.876297  & 0.883079  &       & \cellcolor[rgb]{ .851,  .851,  .851}0.806105  & 0.795215  & 0.792692  &       & \cellcolor[rgb]{ .851,  .851,  .851}0.807842  & 0.797185  & 0.753683  \\
          &       & \cellcolor[rgb]{ .851,  .851,  .851}0.887575  & 0.874909  & 0.881305  &       & \cellcolor[rgb]{ .851,  .851,  .851}0.803610  & 0.792900  & 0.790693  &       & 0.761376  & \cellcolor[rgb]{ .851,  .851,  .851}0.764723  & 0.747315  \\
\cmidrule{2-5}\cmidrule{7-9}\cmidrule{11-13}          & \multirow{3}[2]{*}{8} & \cellcolor[rgb]{ .851,  .851,  .851}0.949686  & 0.920763  & 0.918742  &       & 0.831137  & \cellcolor[rgb]{ .851,  .851,  .851}0.876929  & 0.803050  &       & 0.832800  & 0.772671  & \cellcolor[rgb]{ .851,  .851,  .851}0.842953  \\
          &       & \cellcolor[rgb]{ .851,  .851,  .851}0.948975  & 0.917584  & 0.910023  &       & 0.828118  & \cellcolor[rgb]{ .851,  .851,  .851}0.845975  & 0.799986  &       & 0.795997  & 0.759369  & \cellcolor[rgb]{ .851,  .851,  .851}0.831775  \\
          &       & \cellcolor[rgb]{ .851,  .851,  .851}0.948174  & 0.906219  & 0.901292  &       & \cellcolor[rgb]{ .851,  .851,  .851}0.824221  & 0.730348  & 0.775434  &       & 0.739838  & 0.689923  & \cellcolor[rgb]{ .851,  .851,  .851}0.765730  \\
\cmidrule{2-5}\cmidrule{7-9}\cmidrule{11-13}          & \multirow{3}[2]{*}{10} & \cellcolor[rgb]{ .851,  .851,  .851}0.977641  & 0.927666  & 0.937582  &       & 0.874819  & \cellcolor[rgb]{ .851,  .851,  .851}0.896317  & 0.841704  &       & 0.841063  & 0.717168  & \cellcolor[rgb]{ .851,  .851,  .851}0.860676  \\
          &       & \cellcolor[rgb]{ .851,  .851,  .851}0.977311  & 0.923441  & 0.902343  &       & \cellcolor[rgb]{ .851,  .851,  .851}0.871324  & 0.844036  & 0.838256  &       & \cellcolor[rgb]{ .851,  .851,  .851}0.780500  & 0.717081  & 0.706632  \\
          &       & \cellcolor[rgb]{ .851,  .851,  .851}0.976506  & 0.917141  & 0.901477  &       & \cellcolor[rgb]{ .851,  .851,  .851}0.867734  & 0.715250  & 0.830394  &       & \cellcolor[rgb]{ .851,  .851,  .851}0.747687  & 0.696061  & 0.686917  \\
    \bottomrule
    \end{tabular}}%
  \label{tab:HVonWFG}%
\end{table*}%

All instances  run 20 times independently and their average running times are listed in Table \ref{tab:AVGRuntimesOnWFG}.
As it can be seen, MOEA/D-LIU takes less time to solve each instance than MOEA/D and MOEA/DD,  except for the 3-objective instance of WFG8, indicating that MOEA/D-LIU is computationally efficient on problems WFG1 to WFG9.

In paper \citep{MOEADD}, the best, median and worst HV values obtained by MOEA/DD, MOEA/D, GrEA and HypE on WFG1 to WFG9 instances with different number of objectives are compared, and MOEA/DD and GrEA win on all of the comparisons.
To verify the performance of MOEA/D-LIU on these problems, the HV values obtained by it are compared with the experimental results appeared in that paper, but only those obtained by MOEA/DD and GrEA are kept for comparisons.
The detailed data are listed in Table \ref{tab:HVonWFG}, which can be concluded as follows.
\begin{itemize}
  \item WFG1: MOEA/DD wins on all of the comparisons of this problem except the worst value of the 3-objective instance, and is considered as the  best optimizer for this problem.
  \item WFG2: MOEA/DD wins on the 8-objective instance. As for other instances of this problem, MOEA/D-LIU and MOEA/DD  have their own advantages, and both of them perform better than GrEA.
  \item WFG3: MOEA/D-LIU shows the best performance on the 3- and 5-objective instances, MOEA/DD performs the best on the 8-objective instance, and MOEA/DD wins on the comparisons of the 10-objective instance.
  \item WFG4: MOEA/D-LIU wins on all of the comparisons except the worst value of the 3-objective instance, and is considered as the best optimizer this problem.
  \item WFG5: Again, MOEA/D-LIU wins on all of the comparisons except the worst value of the 3-objective instance, and is also considered as the best optimizer this problem.
  \item WFG6: MOEA/D-LIU wins on all of the comparisons except the best value of the 10-objective instance , and is considered as the best optimizer this problem.
  \item WFG7: MOEA/D-LIU wins on all of the comparisons, and is considered as the best optimizer for this problem.
  \item WFG8: Although MOEA/DD wins on the best and median HV values of the 8-objective instance, and the best value of the 10-objective instance, MOEA/D-LIU wins on the remaining 9 comparisons. In addition, MOEA/D-LIU seems to be more stable on the 8- and 10-objective instances since the difference between the best, median and worst values obtained by MOEA/D-LIU on these two instances is not very significant compared to the values obtained by MOEA/DD.
  \item WFG9: GrEA shows the best performance on the 8-objective instance, and the three algorithms have their own advantages on the other instances of this problem.
\end{itemize}

On the whole, MOEA/D-LIU wins on 69 out of 108 comparisons, and shows very competitive performance on these problems, especially on WFG4 to WFG8.

\subsection{Parameter Sensitivity Studies}\label{parameterSensitivityStudies}
\begin{figure*}[!htbp]
\begin{minipage}[t]{0.24\linewidth}
    \centering
    \includegraphics[width=\textwidth]{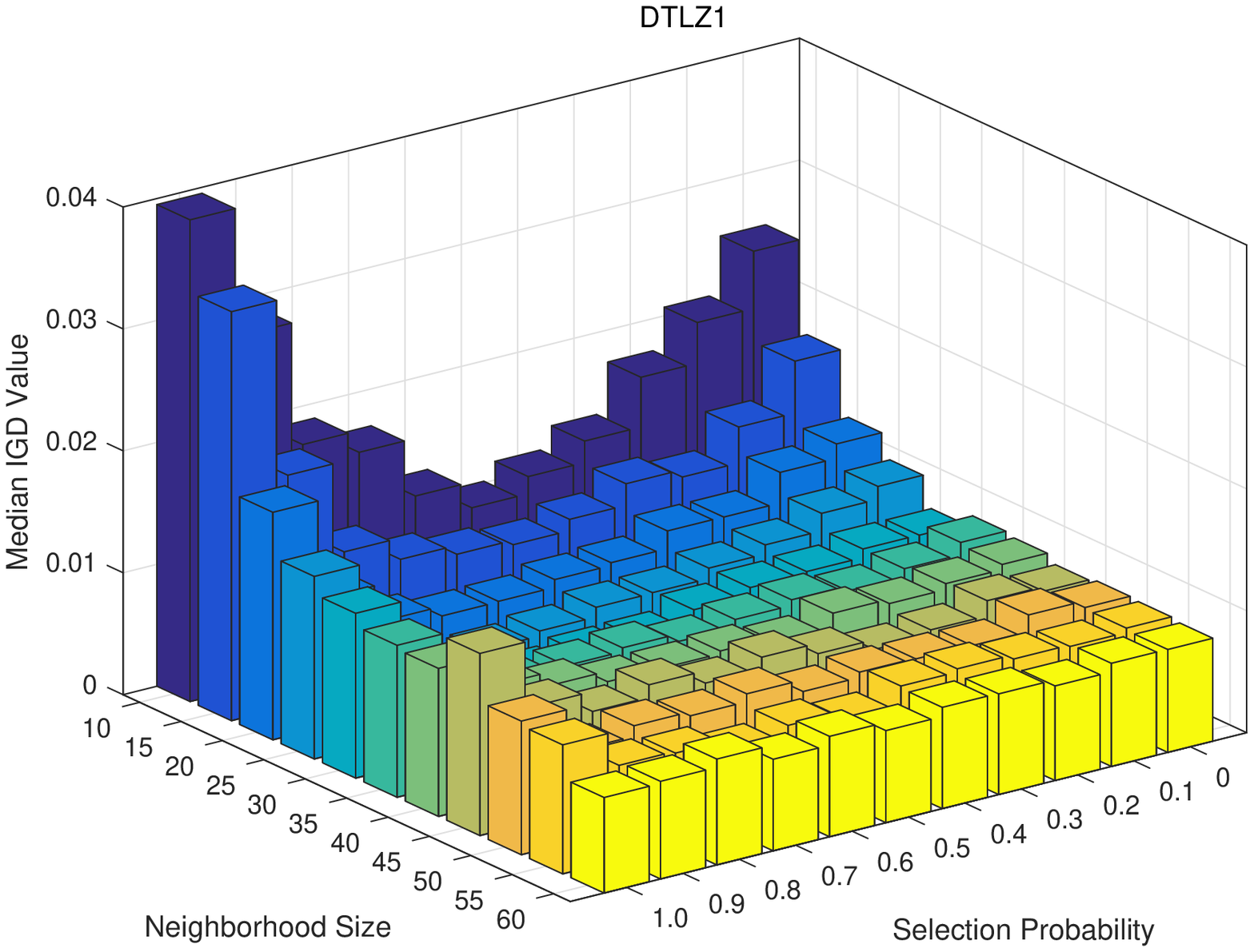}
\end{minipage}
\begin{minipage}[t]{0.24\linewidth}
    \centering
    \includegraphics[width=\textwidth]{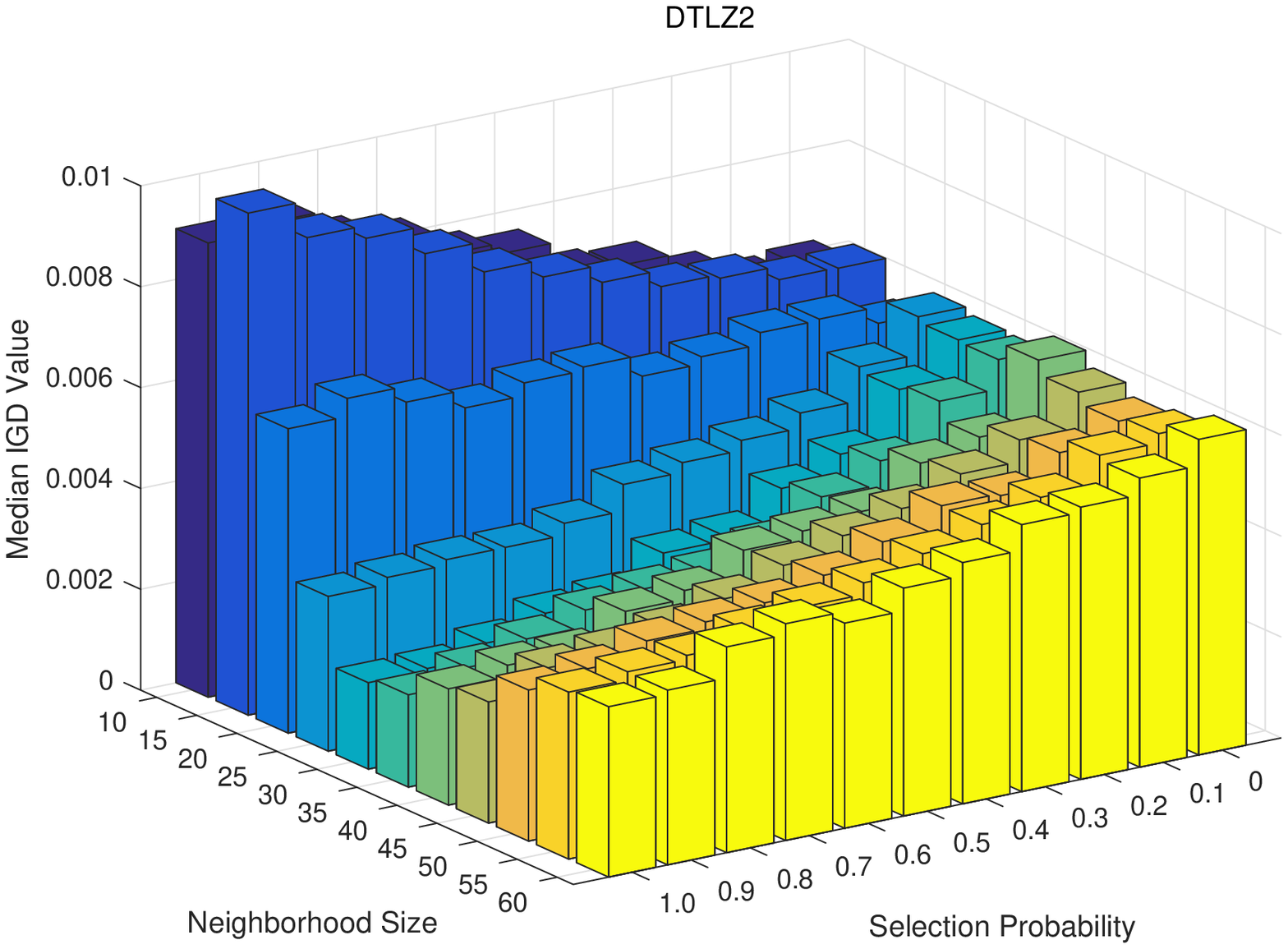}
\end{minipage}
\begin{minipage}[t]{0.24\linewidth}
    \centering
    \includegraphics[width=\textwidth]{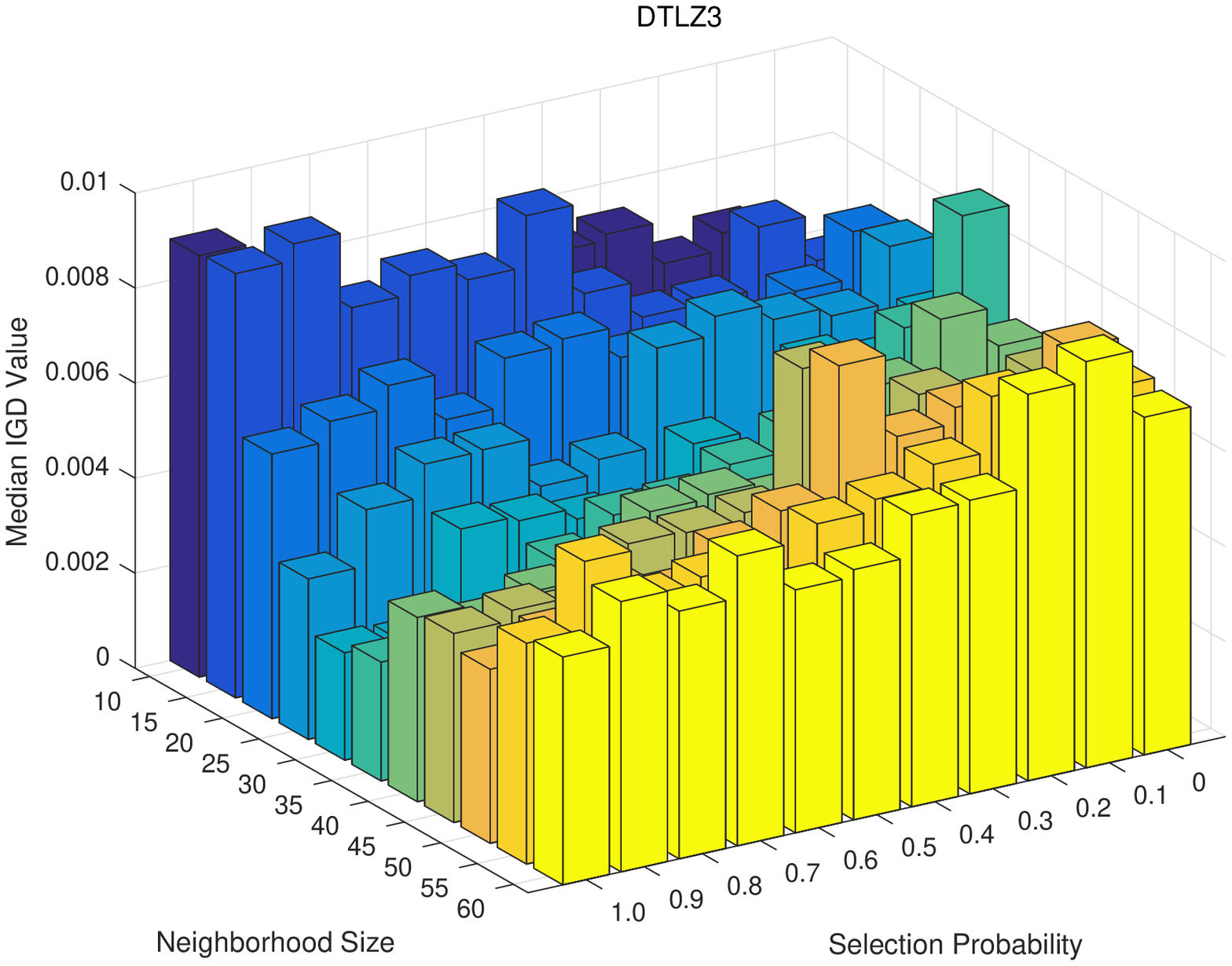}
\end{minipage}
\begin{minipage}[t]{0.24\linewidth}
    \centering
    \includegraphics[width=\textwidth]{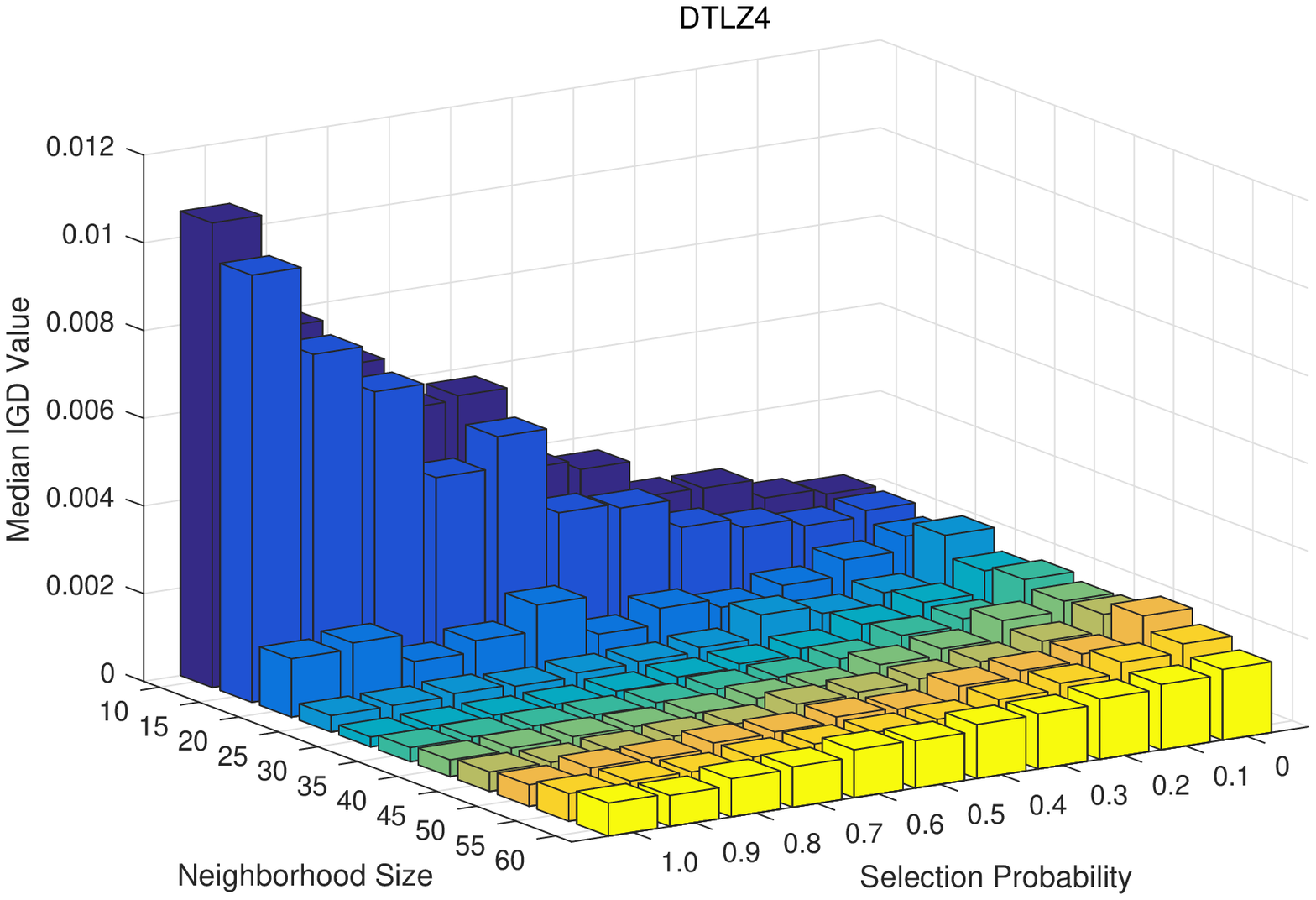}
\end{minipage}
\caption{Median IGD values obtained by MOEA/D-LIU with $p_s$ varying from 0 to 1 with a step size of 0.1, and T taking values 10 to 60 with a step size of 5, on the 15-objective instance of DTLZ1, DTLZ2, DTLZ3 and DTLZ4, respectively.}
\label{fig:T-P-DTLZ}
\end{figure*}
MOEA/D-LIU adopts  a mating selection operation that is often used in the decomposition-based MOEAs. That is, it selects mating parents from the neighborhood of the current weight vector with a high probability $p_s$, where the neighborhood size T plays a key role.
The question is that whether such a neighborhood selection operation is necessary for MOEA/D-LIU to obtain better performance.
It is clear that the neighborhood selection and global selection\footnote{Global selection means always selecting mating parents from the entire population randomly.} would have the same effect, if the individuals in the population are organized randomly and there is no similarity between solutions of neighboring subproblems.
This question can hence be reinterpreted as whether keeping the similarity of solutions of neighboring subproblems can help to explore local areas more effectively.
To answer this question, eleven values are considered for T: varying from 10 to 60 with a step size of 5; eleven values are considered for $p_s$: ranging from 0 to 1 with a step size of 0.1.

The 15-objective instances of DTLZ1 to DTLZ4 are taken as the test instances, and twenty independent runs have been conducted on each test instance to observe the performance of MOEA/D-LIU.
All parameters except the neighborhood size T and the selection probability $p_s$ are the same as previously set in Section \ref{parameterSettings}.
Fig. \ref{fig:T-P-DTLZ} is the histograms of the median IGD values obtained by the algorithm.

Notice that, $p_s$=0 means always selecting mating parents randomly from the entire population. As is can be seen from Fig. \ref{fig:T-P-DTLZ}, such settings do not allow MOEA/D-LIU to achieve better performance than other reasonable settings, implying that the neighborhood selection operation is necessary. It also reflects that keeping the similarity of solutions of neighboring subproblems can help to explore local areas more effectively, and make full use of the advantages of decomposition as mentioned in Section \ref{discussion}.

Other unreasonable settings of T and P can also lead to bad performance of the algorithm,
especially when T and P take a small value and a large value, respectively.
For example, the algorithm show bad performance on all of the instances when T=10 and $p_s$=1.
This is because a small value of T and a large value of P make the reproduction operation excessively explore local areas.

As it can be observed, the algorithm shows good performance when T is between 25 or 40, and the performance gets better and better as $p_s$ goes up, except the case that $p_s=1.0$ for DTLZ1.
Therefore, it is generally better to set T between 25 and 40, and $p_s$ between 0.7 and 0.9.


\section{Conclusion}\label{secConclusion}
In this paper, we propose a simple decomposition-based MOEA with the so-called LIU strategy, i.e., MOEA/D-LIU.
The main ideas of MOEA/D-LIU can be concluded as follows.
Firstly, MOEA/D-LIU employs a set of weight vectors to decompose a given MaOP into a set of subproblems
and optimizes them simultaneously, which is similar to other decomposition-based MOEAs.
Secondly, each weight vector owns only one slot to keep the best individual found so far from the start of the algorithm. This is different from MOEA/DD, in which each weight vector determines a subregion that can have zero, one or more individuals.
Thirdly, unlike MOEA/D, only the worst solution in the current neighborhood will be swapped out and abandoned at each update in MOEA/D-LIU, which can effectively prevent the current population from being occupied by copies of a few individuals, and losing its diversity.
Finally,  the LIU strategy is designed not only to preserve better solutions to the next generation, but also to assign the right solution to each subproblem. This helps the neighborhood selection operation to explore local areas more effectively.

MOEA/D-LIU is compared with  several other famous MOEAs on the 3-, 5-, 8-, 10- and 15-objective instances of problems DTLZ1 to DTLZ4, and the 3-, 5-, 8- and 10-objective instances of problems WFG1 to WFG9.
Experimental results show that our algorithm wins on most of the comparisons and be able to find a well-converged and well-distributed set of solutions.
In addition, although  the time complexity of our algorithm is the same as that of MOEA/D,
the actual running times of MOEA/D-LIU on almost all test instances of problems DTLZ1 to DTLZ4 and WFG1 to WFG9 are less than those of MOEA/D, indicating that our algorithm is computationally efficient.

Our future work can be carried out in the following two aspects.
On the one hand, it is interesting to study the performance of MOEA/D-LIU on other MaOPs,
such as the combinatorial optimization problems appeared in\citep{Zitzler1999Multiobjective,Ishibuchi2010Many}.
On the other hand, it is necessary to improve MOEA/D-LIU to overcome its potential shortcomings.
Especially, the experimental results of our algorithm on DTLZ1 and WFG1 are worse than those of MOEA/DD.




\begin{thebibliography}{10}
\expandafter\ifx\csname url\endcsname\relax
  \def\url#1{\texttt{#1}}\fi
\expandafter\ifx\csname urlprefix\endcsname\relax\def\urlprefix{URL }\fi
\expandafter\ifx\csname href\endcsname\relax
  \def\href#1#2{#2} \def\path#1{#1}\fi

\bibitem{PF}
K.~Miettinen, Nonlinear Multiobjective Optimization, Norwell, MA:Kluwer, 1999.

\bibitem{Jin2001Adapting}
Y.~Jin, T.~Okabe, B.~Sendho, Adapting Weighted Aggregation for Multiobjective
  Evolution Strategies, Springer Berlin Heidelberg, 2001.

\bibitem{Jaszkiewicz2002On}
A.~Jaszkiewicz, On the performance of multiple-objective genetic local search
  on the 0/1 knapsack problem - a comparative experiment, Evolutionary
  Computation IEEE Transactions on 6~(4) (2002) 402--412.

\bibitem{Ishibuchi1998}
H.~Ishibuchi, T.~Murata, A multi-objective genetic local search algorithm and
  its application to flowshop scheduling, IEEE Transactions on Systems, Man,
  and Cybernetics, Part C (Applications and Reviews) 28~(3) (1998) 392--403.
\newblock \href {http://dx.doi.org/10.1109/5326.704576}
  {\path{doi:10.1109/5326.704576}}.

\bibitem{MOEAD}
Q.~Zhang, H.~Li, Moea/d: A multiobjective evolutionary algorithm based on
  decomposition, IEEE Transactions on Evolutionary Computation 11~(6) (2007)
  712--731.
\newblock \href {http://dx.doi.org/10.1109/TEVC.2007.892759}
  {\path{doi:10.1109/TEVC.2007.892759}}.

\bibitem{Tam2016}
H.~H. Tam, M.~F. Leung, Z.~Wang, S.~C. Ng, C.~C. Cheung, A.~K. Lui, Improved
  adaptive global replacement scheme for moea/d-agr, in: 2016 IEEE Congress on
  Evolutionary Computation (CEC), 2016, pp. 2153--2160.
\newblock \href {http://dx.doi.org/10.1109/CEC.2016.7744054}
  {\path{doi:10.1109/CEC.2016.7744054}}.

\bibitem{RVEA}
R.~Cheng, Y.~Jin, M.~Olhofer, B.~Sendhoff, A reference vector guided
  evolutionary algorithm for many-objective optimization, IEEE Transactions on
  Evolutionary Computation 20~(5) (2016) 773--791.
\newblock \href {http://dx.doi.org/10.1109/TEVC.2016.2519378}
  {\path{doi:10.1109/TEVC.2016.2519378}}.

\bibitem{Chen2017}
J.~Chen, J.~Li, B.~Xin, Dmoea- $varepsilon text{C}$ : Decomposition-based
  multiobjective evolutionary algorithm with the $varepsilon $ -constraint
  framework, IEEE Transactions on Evolutionary Computation 21~(5) (2017)
  714--730.
\newblock \href {http://dx.doi.org/10.1109/TEVC.2017.2671462}
  {\path{doi:10.1109/TEVC.2017.2671462}}.

\bibitem{IDBEA}
M.~Asafuddoula, T.~Ray, R.~Sarker, A decomposition-based evolutionary algorithm
  for many objective optimization, IEEE Transactions on Evolutionary
  Computation 19~(3) (2015) 445--460.
\newblock \href {http://dx.doi.org/10.1109/TEVC.2014.2339823}
  {\path{doi:10.1109/TEVC.2014.2339823}}.

\bibitem{MOEADD}
K.~Li, K.~Deb, Q.~Zhang, S.~Kwong, An evolutionary many-objective optimization
  algorithm based on dominance and decomposition, IEEE Transactions on
  Evolutionary Computation 19~(5) (2015) 694--716.
\newblock \href {http://dx.doi.org/10.1109/TEVC.2014.2373386}
  {\path{doi:10.1109/TEVC.2014.2373386}}.

\bibitem{NSGAIII}
K.~Deb, H.~Jain, An evolutionary many-objective optimization algorithm using
  reference-point-based nondominated sorting approach, part i: Solving problems
  with box constraints, IEEE Transactions on Evolutionary Computation 18~(4)
  (2014) 577--601.
\newblock \href {http://dx.doi.org/10.1109/TEVC.2013.2281535}
  {\path{doi:10.1109/TEVC.2013.2281535}}.

\bibitem{GrEA}
S.~Yang, M.~Li, X.~Liu, J.~Zheng, A grid-based evolutionary algorithm for
  many-objective optimization, IEEE Transactions on Evolutionary Computation
  17~(5) (2013) 721--736.
\newblock \href {http://dx.doi.org/10.1109/TEVC.2012.2227145}
  {\path{doi:10.1109/TEVC.2012.2227145}}.

\bibitem{HYPE}
J.~Bader, E.~Zitzler, Hype: An algorithm for fast hypervolume-based
  many-objective optimization, Evolutionary Computation 19~(1) (2011) 45--76.
\newblock \href {http://dx.doi.org/10.1162/EVCO\_a\_00009}
  {\path{doi:10.1162/EVCO\_a\_00009}}.

\bibitem{MOEAD-DE}
H.~Li, Q.~Zhang, Multiobjective optimization problems with complicated pareto
  sets, moea/d and nsga-ii, IEEE Transactions on Evolutionary Computation
  13~(2) (2009) 284--302.

\bibitem{SystematicApproach}
I.~Das, J.~E. Dennis, Normal-boundary intersection: A new method for generating
  the pareto surface in nonlinear multicriteria optimization problems, Siam
  Journal on Optimization 8~(3) (2006) 631--657.

\bibitem{Cvetkovic2002MOP}
D.~Cvetkovic, I.~C. Parmee, Preferences and their application in evolutionary
  multiobjective optimization, Evolutionary Computation IEEE Transactions on
  6~(1) (2002) 42--57.

\bibitem{Osiadacz1989MOP}
A.~J. Osiadacz, Multiple criteria optimization; theory, computation, and
  application, Optimal Control Applications \& Methods 10~(1) (1989) 89¨C90.

\bibitem{Coit1998Genetic}
D.~Coit, Genetic algorithms and engineering design, Engineering Economist
  43~(4) (1998) 379--381.

\bibitem{SBX}
K.~Deb, R.~B. Agrawal, Simulated binary crossover for continuous search space
  9~(3) (2000) 115--148.

\bibitem{PM}
K.~Deb, M.~Goyal, A combined genetic adaptive search (geneas) for engineering
  design, 1999, pp. 30--45.

\bibitem{jMetal2011}
J.~J. Durillo, A.~J. Nebro, jmetal: A java framework for multi-objective
  optimization, Advances in Engineering Software 42 (2011) 760--771.
\newblock \href {http://dx.doi.org/DOI: 10.1016/j.advengsoft.2011.05.014}
  {\path{doi:DOI: 10.1016/j.advengsoft.2011.05.014}}.

\bibitem{jMetal2015}
A.~J. Nebro, J.~J. Durillo, M.~Vergne, Redesigning the jmetal multi-objective
  optimization framework, in: Proceedings of the Companion Publication of the
  2015 Annual Conference on Genetic and Evolutionary Computation, GECCO
  Companion '15, ACM, New York, NY, USA, 2015, pp. 1093--1100.
\newblock \href {http://dx.doi.org/10.1145/2739482.2768462}
  {\path{doi:10.1145/2739482.2768462}}.

\bibitem{IGD}
P.~A.~N. Bosman, D.~Thierens, The balance between proximity and diversity in
  multiobjective evolutionary algorithms, IEEE Transactions on Evolutionary
  Computation 7~(2) (2003) 174--188.
\newblock \href {http://dx.doi.org/10.1109/TEVC.2003.810761}
  {\path{doi:10.1109/TEVC.2003.810761}}.

\bibitem{HV}
E.~Zitzler, L.~Thiele, Multiobjective evolutionary algorithms: a comparative
  case study and the strength pareto approach, IEEE Transactions on
  Evolutionary Computation 3~(4) (1999) 257--271.
\newblock \href {http://dx.doi.org/10.1109/4235.797969}
  {\path{doi:10.1109/4235.797969}}.

\bibitem{WFGalgorithm}
L.~While, L.~Bradstreet, L.~Barone, A fast way of calculating exact
  hypervolumes, IEEE Transactions on Evolutionary Computation 16~(1) (2012)
  86--95.
\newblock \href {http://dx.doi.org/10.1109/TEVC.2010.2077298}
  {\path{doi:10.1109/TEVC.2010.2077298}}.

\bibitem{DTLZ}
K.~Deb, L.~Thiele, M.~Laumanns, E.~Zitzler, Scalable test problems for
  evolutionary multiobjective optimization (2001) 105--145.

\bibitem{WFGProblems}
S.~Huband, L.~Barone, L.~While, P.~Hingston, A scalable multi-objective test
  problem toolkit, Lecture Notes in Computer Science 3410 (2005) 280--295.

\bibitem{WFG}
S.~Huband, P.~Hingston, L.~Barone, L.~While, A review of multiobjective test
  problems and a scalable test problem toolkit, IEEE Transactions on
  Evolutionary Computation 10~(5) (2006) 477--506.
\newblock \href {http://dx.doi.org/10.1109/TEVC.2005.861417}
  {\path{doi:10.1109/TEVC.2005.861417}}.

\bibitem{Zitzler1999Multiobjective}
E.~Zitzler, L.~Thiele, Multiobjective evolutionary algorithms: a comparative
  case study and the strength pareto approach, IEEE Transactions on
  Evolutionary Computation 3~(4) (1999) 257--271.

\bibitem{Ishibuchi2010Many}
H.~Ishibuchi, Y.~Hitotsuyanagi, N.~Tsukamoto, Y.~Nojima, Many-objective test
  problems to visually examine the behavior of multiobjective evolution in a
  decision space, in: International Conference on Parallel Problem Solving From
  Nature, 2010, pp. 91--100.

\end{thebibliography}

\bibliographystyle{elsarticle-num}

\end{document}